\documentclass[10pt,journal,compsoc]{IEEEtran}
\usepackage{graphicx}
\usepackage{amsmath}
\usepackage{amssymb}
\usepackage{booktabs}
\usepackage{arydshln} 
\usepackage{epsfig}
\usepackage{makecell,rotating}
\usepackage{array}
\usepackage{multirow}
\usepackage{algorithm}  
\usepackage{algpseudocode}

\usepackage[pagebackref,breaklinks,colorlinks]{hyperref}
\usepackage{fancyvrb}
\usepackage{url}

\ifCLASSOPTIONcompsoc

  \usepackage[nocompress]{cite}
\else
  \usepackage{cite}
\fi

\ifCLASSINFOpdf

\else

\fi

\begin{document}

\title{Inherit with Distillation and Evolve with Contrast:\\ Exploring Class Incremental Semantic Segmentation without Exemplar Memory}

\author{Danpei~Zhao*,~\IEEEmembership{Member,~IEEE,}
        Bo~Yuan,
        Zhenwei~Shi,~\IEEEmembership{Member,~IEEE}
\IEEEcompsocitemizethanks{\IEEEcompsocthanksitem Danpei Zhao, Bo Yuan and Zhenwei Shi are with the Image Processing Center, School of Astronautics, Beihang University, Beijing 100191, China.\protect\\
E-mail: \{zhaodanpei, yuanbobuaa, shizhenwei\}@buaa.edu.cn. * Corresponding author.
}
\thanks{$\copyright$2023 IEEE. Personal use of this material is permitted. Permission from IEEE must be obtained for all other uses, in any current or future media, including reprinting/republishing this material for advertising or promotional purposes, creating new collective works, for resale or redistribution to servers or lists, or reuse of any copyrighted component of this work in other works. }}

\markboth{Journal of \LaTeX\ Class Files,~Vol.~x, No.~x, x~x}%
{Shell \MakeLowercase{\textit{et al.}}: Bare Demo of IEEEtran.cls for Computer Society Journals}

\IEEEtitleabstractindextext{%
\begin{abstract}
As a front-burner problem in incremental learning, class incremental semantic segmentation (CISS) is plagued by catastrophic forgetting and semantic drift. Although recent methods have utilized knowledge distillation to transfer knowledge from the old model, they are still unable to avoid pixel confusion, which results in severe misclassification after incremental steps due to the lack of annotations for past and future classes. Meanwhile data-replay-based approaches suffer from storage burdens and privacy concerns. In this paper, we propose to address CISS without exemplar memory and resolve catastrophic forgetting as well as semantic drift synchronously. We present Inherit with Distillation and Evolve with Contrast (IDEC), which consists of a Dense Knowledge Distillation on all Aspects (DADA) manner and an Asymmetric Region-wise Contrastive Learning (ARCL) module.  Driven by the devised dynamic class-specific pseudo-labelling strategy,  DADA distils intermediate-layer features and output-logits collaboratively with more emphasis on semantic-invariant knowledge inheritance. ARCL implements region-wise contrastive learning in the latent space to resolve semantic drift among known classes, current classes, and unknown classes. We demonstrate the effectiveness of our method on multiple CISS tasks by state-of-the-art performance, including Pascal VOC 2012, ADE20K and ISPRS datasets. Our method also shows superior anti-forgetting ability, particularly in multi-step CISS tasks. 
\end{abstract}

}

\maketitle

\IEEEdisplaynontitleabstractindextext

\IEEEpeerreviewmaketitle

\IEEEraisesectionheading{\section{Introduction}\label{Sec-Introduction}}

\IEEEPARstart{I}{ncremental} learning (IL), also known as continual learning, aims to learn a sequence of tasks and expects that it can achieve proper performance in both old and new tasks. Semantic segmentation assigns a label to every pixel in the image. Typically, the popular fully-supervised semantic segmentation methods require large-scale annotations to support model training. 
However, these models are typically designed for a closed set, meaning that they can only handle a fixed number of classes, and all the data must be fed to the model at once. In realistic scenes, data is usually accessed incrementally. Apparently, discarding the obtained models and re-training new ones on new data signifies a waste of time and computing resources. For example, ChatGPT costs ~\$4.5 million for one-time training. And sometimes the old data can not be accessible due to privacy restrictions. On the other side, simply re-training the model will bring an Alzheimer-like problem, i.e., the model will lose the past ability due to the parameter update~\cite{LWF}. In this case, class incremental semantic segmentation (CISS) is a promising but challenging task that is relevant to practical vision computing fields such as remote-sensing observation and automatic driving, etc. 
\begin{figure}[t]
	\centering
	\includegraphics[scale=0.73]{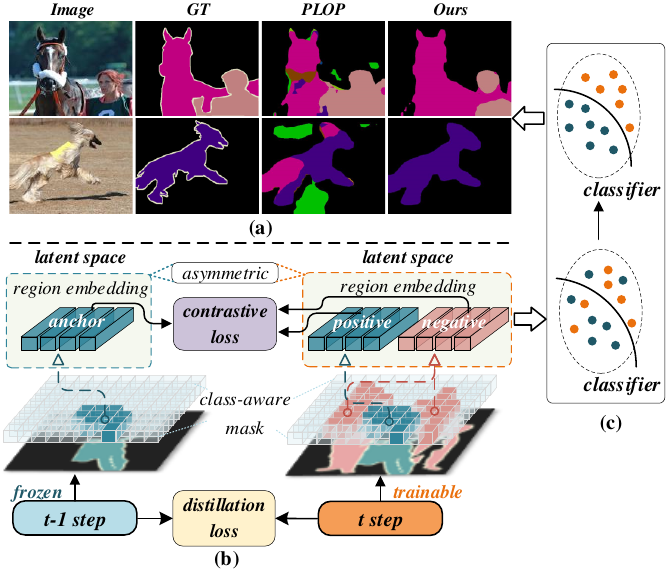}
	\caption{Schematic illustration of the proposed approach. (a): Visualization comparison of PLOP~\cite{PLOP} and the proposed IDEC. (b): Main structure of IDEC at $t$ step training. (c): Sketchy demonstration of classifier evolution in IDEC. The network can be trained in an end-to-end way.}
	\label{fig-motivation}
\end{figure}

In recent years, IL in deep learning has been exploited in many visual tasks~\cite{OCF, DeLange2021ACL}. The main obstacle of IL is \emph{catastrophic forgetting}~\cite{ramasesh2022effect}, i.e., models fail to update the parameters for learning new classes meanwhile preserving sufficient performance on the old ones. In terms of downstream performance, models often encounter classifier bias and pixel misclassification. 
As is well-known, a common strategy to alleviate catastrophic forgetting is knowledge distillation (KD)~\cite{Hinton2015DistillingTK}. It normally follows a Teacher-Student architecture. KD has been proven to be effective in mitigating catastrophic forgetting in image classification~\cite{EWC, iCaRL, ABD, Zhu2021PrototypeAA, Liu2021AdaptiveAN, Zhang2021FewShotIL, Rostami2021LifelongDA, Tang2022LearningTI, Zhou2022ForwardCF} and object detection~\cite{Shmelkov2017IncrementalLO, Joseph2021IncrementalOD, Liu2020MultitaskIL, Wang2021WanderlustOC, Feng2022OvercomingCF}. However, CISS is more challenging due to the dense prediction demand and complex context association. 

Besides catastrophic forgetting, CISS encounters another challenge beyond the image classification task, which is the \emph{semantic drift}. As illustrated in Fig.~\ref{fig-motivation}(a), due to the lack of old data, the models tend to confront class confusion and classifier bias. In addition, since only the current classes are labelled at each incremental step, the semantic of background pixels drifts because the connotation varies, i.e., known classes and future classes are mixed as one \emph{background} class. As a consequence, it will lead to subsequent classification chaos. Some recent works~\cite{MiB, SSUL} try modelling the unknown background class by specifying classifier initialization or taking advantage of visual saliency. While existing exemplar-free approaches like~\cite{MiB, PLOP, RCIL} concentrate on the consistency of the endmost outputs between the old model and the current model. However, these pioneers fail to preserve the model structure consisting of internal feature distribution, causing the classifier bias on new classes, especially in multi-step IL tasks. In this paper, we present Inherit with Distillation and Evolve with Contrast (IDEC), addressing the CISS problem from two mutually reinforced manners performance in the latent space to maintain inner feature consistency and solid knowledge transfer.  

As depicted in Fig.~\ref{fig-motivation}(b), at step \emph{t}, the \emph{t-1} step model is frozen to generate the predictions as the pseudo label for \emph{t} step training. 
Specifically, we propose to resolve CISS from two aspects. On the one hand, we utilize knowledge distillation to inherit from the old model and a customized distillation loss is introduced between the old and current models. Superior to the previous KD-based methods~\cite{PLOP, ILT}, our distillation manner considers every intermediate layer and output logits synchronously with more emphasis on semantic-invariant knowledge inheritance. Motivated by human recognition patterns~\cite{Treisman1980AFT}, we argue that the model tends to recognize and memorize things through semantic-invariant parts. For solid knowledge transfer, we believe the semantic-invariant term occupies a dominant position. Thus in our distillation strategy, we put more emphasis on the deep layers, which contain more semantic-invariant term. While for the low layers, the representation of detail features serves as the variance part to contribute to the intra-class variance. Considering different emphases on low- and high-level features in semantic interpretation, we design an attenuated layer-aware weight (ALW) to guide the distillation process. On the other hand, we pay attention to the adaptation on new classes in a specific contrastive learning manner. Obviously, the latent space of the old model and the current model is asymmetric since the cluster numbers are unequal since the additional classes at incremental steps, as the classifier varies (the number of prediction classes is changed). We run contrastive learning in this asymmetric latent space by selecting specific \textit{anchor} from the old model, \textit{positive} and \textit{negative} embeddings from the current model, to alleviate the semantic drift and classifier bias.

Our contributions are summarized as follows.
\begin{itemize}
\item[$\bullet$] We propose to address catastrophic forgetting and semantic drift synergistically for CISS in an end-to-end way. And we propose two mutually reinforced modules to enhance the anti-forgetting ability on old classes and compatibility with new classes synchronously.
\item[$\bullet$] We present an efficient dense distillation strategy and an asymmetric region-wise contrastive learning mechanism in the latent space without requiring exemplar memory.
\item[$\bullet$] We devise a class-specific pseudo-labelling strategy by launching a dynamic threshold to boost the continual updating with self-training. 
\item[$\bullet$] The proposed approach achieves state-of-the-art performance in large-scale datasets including Pascal VOC 2012, ADE20K and ISPRS, and shows advantages in challenging multi-step CISS tasks.
\end{itemize}

\section{Related Work}
In this section, we review the previous researches on IL, CISS, KD and contrastive learning, respectively. By summarizing and analyzing the defects of the current cutting-edge, we propose to resolve catastrophic forgetting and semantic drift synergistically.

\subsection{Class Incremental Learning}
Incremental learning (IL) technique breaks through the typical one-off training process in deep learning. It enables a neural network to continually update its parameters to adjust incremental data. It has been explored in computer vision~\cite{Lomonaco2022CVPR2C, Qu2021RecentAO, Belouadah2021ACS}, natural language processing~\cite{Autume2019EpisodicMI, Biesialska2020ContinualLL} and remote sensing~\cite{Bhat2021CILEANETCI}. The most challenge of this technique is catastrophic forgetting due to the parameter updating. The problem of catastrophic forgetting has been discovered and discussed as early as the 1980s by McCloskey, \textit{et al.}~\cite{McCloskey1989CatastrophicII}. That is algorithms trained with backpropagation suffers from severe knowledge forgetting just like human suffers from gradual forgetting of previously learned tasks. 
To alleviate this problem, a range of researches~\cite{RainbowMC, IL2MCI, UsingHT, ContinualLO, OnlineCL} propose to retrospect known knowledge including sample selection as exemplar memory~\cite{Aljundi2019GradientBS, Han2018CoteachingRT, Rolnick2019ExperienceRF, Fini2020OnlineCL, Shi2021ContinualLV}, prototypes guidance~\cite{Ho2021PrototypesGuidedMR, Ahn2019UncertaintybasedCL, Zhu2021PrototypeAA, Zhu2021SelfPromotedPR}, meta learning~\cite{Javed2019MetaLearningRF, Wang2020EfficientML, Banayeeanzade2021GenerativeVD, Hurtado2021OptimizingRK}, generative adversarial learning~\cite{Ebrahimi2020AdversarialCL, Xiang2019IncrementalLU, Verma2021EfficientFT} and so on. These approaches can achieve effective continual learning performances but normally require extra memory to store old data. However, a more challenging but practical IL scene indicates no old data is available. In recent years, some researches concentrate on IL without data replay by using knowledge distillation~\cite{Kang2022ClassIncrementalLB, Douillard2020PODNetPO, Ye2020DistillingCK, Tian2021RelationshipPreservingKD, Xu2022DeepNN, Zhao2022EmbeddedSI}, conducting weight transfer~\cite{Ahn2019VariationalID, Slim2022DatasetKT} and network architecture extension~\cite{Kanakis2020ReparameterizingCF, Liu2021AdaptiveAN, Yan2021DERDE}.  The present primary concern in IL is to balance the anti-forgetting and new knowledge assimilating, which is the priority in this paper.

\subsection{Class Incremental Semantic Segmentation}
Current class incremental learning (CIL) concentrates on image classification~\cite{LWF, Liu2021AdaptiveAN, serra2018overcoming, Zhu2021PrototypeAA, Zhou2022ForwardCF}. Some recent approaches~\cite{MiB, PLOP} extend it to more challenging dense prediction tasks. Class incremental semantic segmentation (CISS) faces two main challenges including catastrophic forgetting and semantic drift, which are because of the lacking of old data and parameter update~\cite{Hu2021DistillingCE, Kaushik2021UnderstandingCF, RW}. According to whether to use exemplar memory, CISS approaches can be divided into two categories.  For the first kind,~\cite{ILT, CIL, PLOP, UCD, RCIL} utilize knowledge distillation to inherit the capability of the old model. Cermelli \textit{et al.}~\cite{MiB} propose to reduce semantic drift by modelling the background label. In remote sensing, researches focus on small objects enhancing~\cite{Li2022ClassIncrementalLN} and multi-level distillation~\cite{Tasar2019IncrementalLF, Shan2022ClassIncrementalLF}. The second kind stores a portion of past training data as exemplar memory. Rebuffi~\textit{et al.}~\cite{iCaRL} propose to conduct the incremental steps with the supervision of representative past training data. However, it will update the old model which may aggravate catastrophic forgetting. Cha~\textit{et al.}~\cite{SSUL} proposes a class-imbalanced sampling strategy and uses a visual saliency detector to filter unknown classes. Maracani \textit{et al.}~\cite{RECALL} resort to relying on a generative adversarial network or web-crawled data to retrieve images. These data-replay-based methods bring extra memory consumption, which is critical especially under CISS circumstances. In recent years, few-shot/zero-shot approaches~\cite{Cheraghian2021SemanticawareKD, Xu2020ProgressiveDF} are also explored to reduce data and annotation dependency at incremental steps. Whereas these methods normally have low performance in complex visual tasks. Based on the analysis above, in this paper we propose to address CISS without exemplar memory, to meet with realistic applications. 

\subsection{Knowledge Distillation}
In deep learning, knowledge distillation (KD) aims to transfer information learned from one model to another whilst training constructively. KD was firstly defined by~\cite{MC} and generalized by~\cite{Hinton2015DistillingTK}. The common characteristic of KD is symbolized by its Student-Teacher (S-T) framework~\cite{Wang2022KnowledgeDA}. In recent years, KD has been applied in model compression~\cite{Furlanello2018BornAN, Yang2020ModelCW}, knowledge transfer~\cite{Yang2019SnapshotDT, Heo2019ACO} and semantic segmentation~\cite{2020Structured}. And for dense prediction tasks, knowledge distillation has been demonstrated feasible and effective~\cite{Wang2022KnowledgeDA}. For example, pixel-wise similarity distillation~\cite{Feng2021DoubleSD} and channel-wise distillation~\cite{Shu2021ChannelwiseKD} are proposed to improve the distillation efficiency. Wang~\textit{et al.}~\cite{Wang2020IntraclassFV} propose to transfer the intra-class feature variation from teacher to student by constructing densely pairwise relations. In IL tasks, Zhou~\textit{et al.}~\cite{Zhou2019M2KDMA} propose a multi-level knowledge distillation strategy by leveraging all previous model snapshots. In CISS, KD has been proven as an effective way to preserve the capability of classifying old classes without storing past data in incremental steps.  A typical KD approach is to use the outputs from the old model to guide the new model by a distillation loss~\cite{MiB}. Further, Michieli~\textit{et al.}~\cite{ILT} explores distillation in intermediate feature space and indicates that L2-norm is superior to cross-entropy or L1-norm. Qiu~\textit{et al.}~\cite{Qiu2022SATSST} use self-attention to capture both within-class and between-class knowledge.

\subsection{Contrastive Learning}
Contrastive learning has been widely used in learning representations without labels~\cite{chen2020improved, hjelm2018learning, Wu2018UnsupervisedFL}. A typical contrastive learning way is to apply diversiform transforms on the source image to create positive pairs~\cite{SimCLR, komodakis2018unsupervised}. For dense prediction tasks, Lee \textit{et al.}~\cite{Lee2021AttentionGuidedSC} propose a supervised contrastive learning approach via highlighting semantic attention. Wang \textit{et al.}~\cite{Wang2021ExploringCP} propose a pixel-wise contrastive algorithm for semantic segmentation in the fully-supervised setting. In recent years, similar to metric learning approaches, some contrastive methods attempt to learn the representations in the latent space by organizing positive embedding pairs against negative embedding pairs~\cite{SDR, Kan2021RelativeOA}.  Arnaudo \textit{et al.}~\cite{Arnaudo2022ACD} explore a contrastive regularization between outputs of the old model and the new model but only rely on image transform. Recent state-of-the-art CISS methods SDR~\cite{SDR} and UCD~\cite{UCD} propose to compare feature vectors in the feature space. 
By contrast, the differences and advantages of the proposed method compared to~\cite{SDR, UCD} are: 1) we perform contrastive learning in asymmetric latent space, enabling to enlarge divergence among the new classes, known classes the future/unknown classes; 2) we construct \emph{anchor} embedding from the old model, which is more reliable by avoiding the impact of model performance degradation in IL steps; 3) our method supports region-wise embedding comparison, superior to the pixel-wise calculation in dense prediction task. By considering intact context association, the proposed model achieves more robust performance in multi-step CISS tasks.

\begin{figure*}[t]
	\centering
	\includegraphics[scale=0.75]{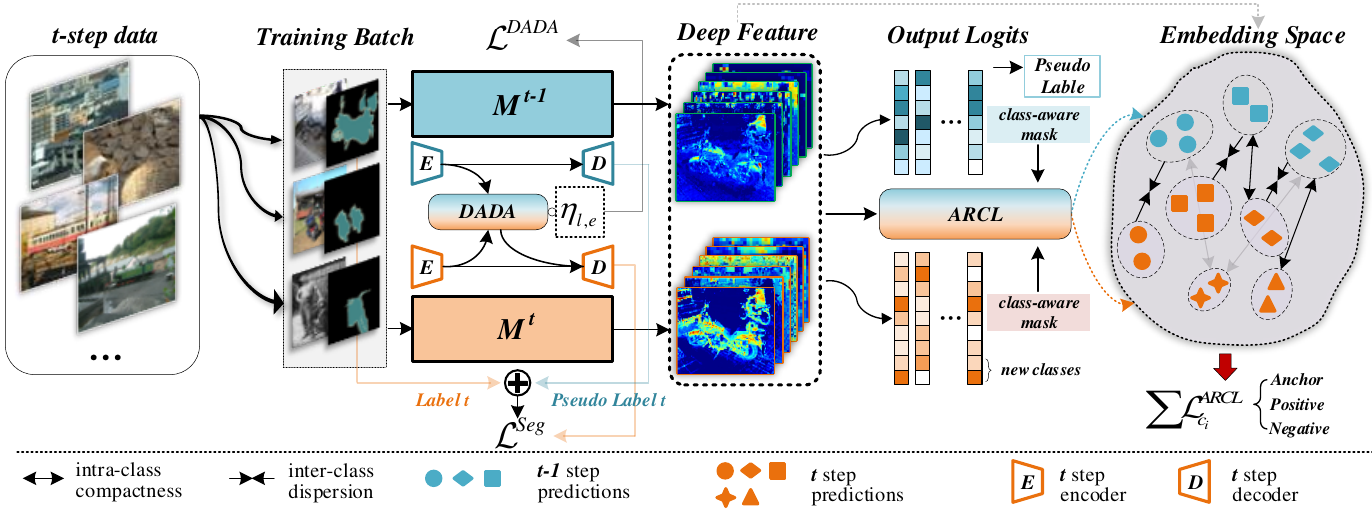}
	\caption{The pipeline of the proposed IDEC at \textit{t} step training: Each training batch is imported to both $M^{t-1}$ and $M^t$. We propose DADA to run distillation within immediate feature space and output space, supervised by a unified distillation loss $\mathcal{L}^{DADA}$. Utilizing output logits from $M^{t-1}$ and $M^t$, we extract corresponding deep features tagged with prediction class to run contrastive learning in the latent space. Abstractly, we pick \textit{anchor} embedding from $M^{t-1}$, select \textit{positive} and \textit{negative} embeddings from $M^t$. A contrastive loss $\mathcal{L}^{ARCL}$ is introduced to supervise the asymmetric proposed region-wise contrastive learning. $\mathcal{L}^{Seg}$ calculates pixel classification loss of  $M^t$ under the supervision of \textit{t} step labels and pseudo labels from \textit{t-1} step.}
	\label{fig-network}
\end{figure*}
\section{Methodology}
\label{Sec-Methodology}
In this section, we present a CISS method named Inherit from Distillation and Evolve with Contrast (IDEC). Fig.~\ref{fig-network} depicts the information flow at \textit{t} step training. The main structure of the proposed network consists of the following subsections.

\subsection{Preliminary}
Let $\mathcal{D}={(x_i, y_i)}$ signifies the training dataset, where $x_i \in \mathbb{R}^{C\times H\times W}$ denotes the input image and $y_i \in \mathbb{R}^{H\times W}$ denotes the corresponding ground truth. $\mathcal{D}^t$ indicates the training dataset for \emph{t} step. At $t$ step, $C^{0:t-1}$ indicates the previously learned classes and $C^t$ indicates the classes for learning. When training on $\mathcal{D}^t$, the training data of old classes, i.e., $\{\mathcal{D}^0, \mathcal{D}^1, \cdots, \mathcal{D}^{t-1}\}$ is inaccessible. And the ground truth in $\mathcal{D}^t$ only covers $C^t$. The complete training process consists of \{Step-0, Step-1, $\cdots$, Step-T\} steps. 
At $t$ step, we use $M^{t-1}$ and $M^t$ to represent the \textit{t-1} and \textit{t} step model, respectively. $\mathcal{F}_l^t, l\in N_l$ indicates the $l$-th layer feature map from $M^t$ generated by feature extractor \textit{F}. $N_l$ is the layer number of the feature extractor network.  

\subsection{Dense Alignment Distillation on All Aspects}
\label{Sec-DADA}
\begin{figure}[h]
	\centering
	\includegraphics[scale=0.9]{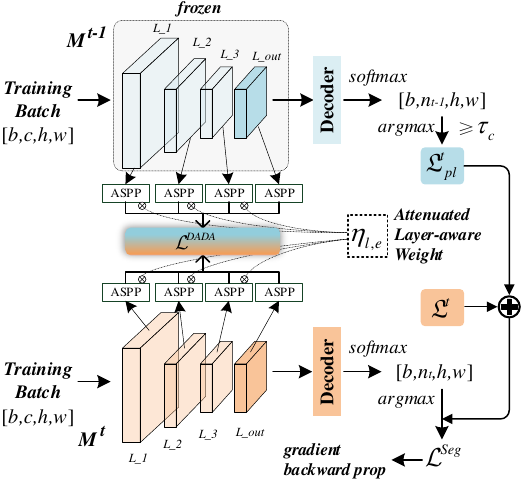}
	\caption{The proposed DADA performs knowledge distillation in every intermediate-layer and the output-logits between $M^{t-1}$ and $M^t$ simultaneously. Note that we omit the input layer. $\eta_{l,e}$ is used to guide the immediate-layer feature distillation.}
	\label{fig-DADA}
\end{figure}
To address the catastrophic forgetting problem, we propose a novel knowledge distillation strategy. However, in contrast to the pioneers~\cite{ILT, 2021Knowledge, PLOP}, we propose Dense Alignment Distillation on all  Aspects (DADA) including intermediate layers as well as output logits synchronously but with different emphases on low- and high-level features. Dense feature distillation in multi-layer is necessary for CISS because semantic segmentation requires each pixel to be classified. We propose matching feature representation at different levels is beneficial for inheriting the feature representation of old models in a dense prediction task. To align with the pioneers, we take DeepLabv3~\cite{DeepLabv3} as an example to present DADA. But the strategy can be embedded expediently into any other segmentation models as discussed in Sec.~\ref{Sec-ISM}. As shown in Fig.~\ref{fig-DADA}, the input training batch is imported to both the $M^{t-1}$ and $M^{t}$ to extract multi-layer feature  $\{\mathcal{F}_l^{t-1}, \mathcal{F}_l^t\}, l\in N^l$. Note that $M^{t-1}$ is frozen to avoid gradient backpropagation but $M^{t}$ is trainable. To take full advantage of context association, for each  $\mathcal{F}_l^t$, we employ a typical atrous spatial pyramid pooling (ASPP)~\cite{DeepLabv3} $\Lambda$ to compute feature responses from different scales of the receptive field. After that, we obtain embeddings $\{\textbf{e}_l^{t-1}, \textbf{e}_l^t\}$ through  $\textbf{e}_l=\Lambda (\mathcal{F}_l)$, where $\textbf{e}_l \in \mathbb{R}^{N\times W'\times H'}$, $N$ represents the feature channels. 
We calculate the similarity of the embeddings $\{\textbf{e}_l^{t-1}, \textbf{e}_l^t\}$ by measuring the Kullback-Leibler (K-L) divergence.
\begin{equation}
	d_{D}\left(\textbf{e}_l^{t-1}, \textbf{e}_l^t \right) = \frac{1}{W'\times H'} \sum_i^{W'\times H'} KL(p_i^{t-1} \lVert p_i^{t}),
\end{equation}
where $p_i^{t-1}$ and $p_i^t$ represent class probabilities of the \textit{i}-th pixel in the softmax output when applying a 1$\times$1 convolution on the embedding $\textbf{e}_l^{t-1}$ and $\textbf{e}_l^t$, respectively. 

For the output layer, we measure the embeddings similarity before the classification layer (e.g. softmax), which can be calculated by $d_{D}\left(\textbf{e}_{out}^{t-1}, \textbf{e}_{out}^{t}\right)$. In our implementation, the size of the deepest features $\{\mathcal{F}_{d}^{t-1}, \mathcal{F}_{d}^{t}\}$ is $\frac{H\times W}{16}$. $M^{t-1}$ and $M^t$ share the same semantic segmentation architecture~\cite{DeepLabv3}. In this case, $M^{t}$ can inherit the capacity of  $M^{t-1}$ in feature space by distilling the immediate-layer and output-layer features synchronously.

In CISS tasks, it is necessary to balance old knowledge inheritance and new knowledge learning. Inspired by~\cite{2020Channel}, the impact of distilling $M^{t-1}$ is not always positive, particularly it will inhibit $M^t$ from learning better in the latter training stage. Thus we give less concern to knowledge distillation with training iteration increases. On the other hand, compared to LocalPOD~\cite{PLOP}, we put different emphases on semantic-invariant features (mainly from deep layers) and sample-specific features (mainly from low layers). We believe the former matters more in IL steps because the old data is inaccessible in our CISS setting. Here we propose an Attenuated Layer-aware Weight (ALW) $\eta_{l,e}$ to guide the dense distillation process:
\begin{equation}
	\eta_{l,e} = \alpha \times log\left(1+\frac{n_l}{N_l}\right) \times \gamma^{n_e/N_e}
\end{equation}
where $\alpha$ is an initial weight of intermediate-layer distillation loss. $n_e$ and $N_e$ represent the $e$-th epoch and total training epochs. $n_l$ indicates $l$-th intermediate layer. And $\gamma$ is a positive constant less than 1.The optimization goal for intermediate feature consistency is defined as:
\begin{equation}
	\mathcal{L}^{IL-D} = \frac{1}{N_l} \sum_l^{N_l} \eta_{l,e} \times d_{D}\left(\textbf{e}_l^{t-1}, \textbf{e}_l^t \right)
\end{equation}
In our implementation, we run above distillation across the middle three intermediate layers (by skipping the input layer). Therewith the distillation goal for the output layer is defined as:
\begin{equation}
	\mathcal{L}^{OL-D} = d_{D}\left(\textbf{e}_{out}^{t-1}, \textbf{e}_{out}^t \right)
\end{equation}
Hence the proposed DADA is supervised by
\begin{equation}
		\mathcal{L}^{DADA} =\mathcal{L}^{IL-D} + \lambda \mathcal{L}^{OL-D}
\end{equation}
where $\lambda$ is a constant to balance the contribution of $\mathcal{L}^{IL-D} $ and $\mathcal{L}^{OL-D}$ in the training process. $\mathcal{L}^{DADA}$ is calculated at every incremental step. 

\subsection{Asymmetric Region-wise Contrastive Learning}
\label{Sec-ARCL}
\begin{figure}[t]
	\includegraphics[scale=0.98]{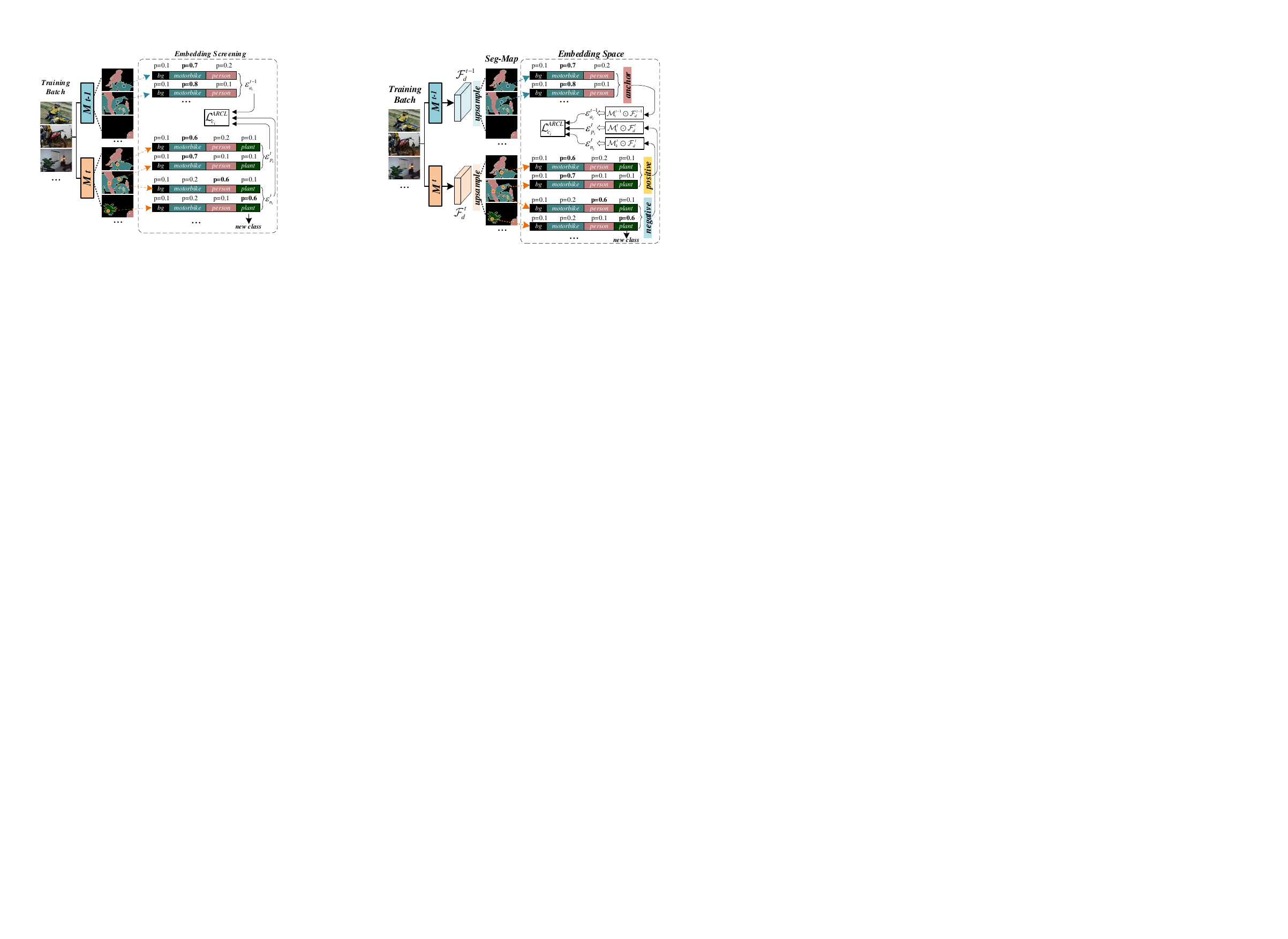}
	\caption{The proposed ARCL which being performed in a training batch. In this example, we tag \textit{motorbike} to select corresponding \textit{anchor} embedding from features of $M^{t-1}$, and select \textit{positive} and \textit{negative} embeddings from $M^t$. \textit{p} indicate the prediction probability belonging to a specific class of the selected pixel. We perform this operation across the whole batch to select all the corresponding region embeddings to calculate the contrastive loss.}
	\label{fig-ARCL}
\end{figure}

At each incremental step, since the lack of annotations, all the other classes are actually labelled as \textit{background} (bg) except for current classes. Thus the connotation of background would contain known classes and future classes simultaneously, leading to semantic drift. 
To alleviate the classifier confusion caused by semantic drift, we propose an asymmetric region-wise contrast learning approach. Specifically, we select \textit{anchor} embedding from the old model since the model faces catastrophic forgetting after IL steps. While the \textit{positive} and \textit{negative} embeddings are generated from the current model. 
Firstly, after obtaining the output logits from $M^{t-1}$ and $M^t$ on $\mathcal{D}^t$, we generate a class-aware mask under the guidance of model prediction. As depicted in Fig.~\ref{fig-ARCL}, we use the output from $\mathcal{F}^{t-1}$ to select \textit{anchor} embeddings for a designated class. The corresponding \textit{positive} and \textit{negative} embeddings are from features $\mathcal{F}^{t}$ generated by $M^t$.  
To achieve this, we first construct the class-aware mask $\mathcal{M}_k$ for class $k$. In detail, $\mathcal{M}_k$ is obtained by:
\begin{equation}
\begin{aligned}
&\mathcal{M}_k = (m_{ij})_{H_d\times W_d} \\
&m_{ij} = \left\{
	\begin{aligned}
	& 1,  \quad \rm{if} \ p_{ij}=k \\ 
	& 0, \ \ \ \rm{otherwise}\\
	\end{aligned}
    \right.   
\end{aligned}
\end{equation}
We perform the embedding selection process in the deepest features $\{\mathcal{F}_{d}^{t-1}, \mathcal{F}_{d}^{t}\}$, where $\mathcal{F}_{d}^{t} \in \mathbb{R}^{N_d\times W_d\times H_d}$. For a designated class $i$, we obtain the \textit{anchor}, \textit{positive} and \textit{negative} embeddings by:
\begin{equation}
	\begin{aligned}
		&\mathcal{E}_{a_i}^{t-1} = \mathcal{M}_i^{t-1} \odot \mathcal{F}_{d}^{t-1} , i \in C^{1:t-1} \\
		&\mathcal{E}_{p_i}^{t} = \mathcal{M}_i^t \odot \mathcal{F}_{d}^{t}, i \in C^{1:t-1} \\
		&\mathcal{E}_{n_i}^{t} = \mathcal{M}_k^t \odot \mathcal{F}_{d}^{t}, i \in C^{1:t-1}, k \neq i\wedge k \in C^{t}
	\end{aligned}
\end{equation}
where $\mathcal{M}_i$ is the mask for filtering $i$-th class embedding in the latent space. $\odot$ indicates the Hadamard product between two matrices. $\mathcal{E}_{a_i}^{t-1}$ represents the \emph{anchor} embedding belonging to $i$-th class in the latent space from $M^{t-1}$. $\mathcal{E}_{p_i}^t$ and $\mathcal{E}_{n_i}^t$ indicate the corresponding \emph{positive} and \emph{negative} embeddings in the latent space from $M^t$. 
After that, for computing the intra-class compactness and inter-class dispersion, we transform the $\mathcal{E}_{a_i}^{t-1}$, $\mathcal{E}_{p_i}^{t}$ and $\mathcal{E}_{n_i}^{t}$ to one-dimensional tensors \{$\hat{\mathcal{E}}_{a_i}^{t-1}$, $\hat{\mathcal{E}}_{p_i}^{t}$, $\hat{\mathcal{E}}_{n_i}^{t}$\}. 
\begin{equation}
		\begin{aligned}
		&\hat{\mathcal{E}}_{a_i}^{t-1}, \hat{\mathcal{E}}_{p_i}^{t}, \hat{\mathcal{E}}_{n_i}^{t} \in \mathbb{R}^{N_d \hat{H}\hat{W}} \\
	\end{aligned}
\end{equation}
where $\hat{H}\hat{W}$ is the minimum length among the size of $\mathcal{E}_{a_i}^{t-1}$, $\mathcal{E}_{p_i}^{t}$ and $\mathcal{E}_{n_i}^{t}$. We perform this embedding screening operation in each training batch to obtain cross-image semantic correlation. In this way, based on contrastive learning across $M^{t-1}$ and $M^t$, DADA and ARCL tackle catastrophic forgetting and semantic drift with a mutually reinforced relationship.
\begin{algorithm}[b]  
	\caption{Pseudocode of the proposed ARCL at $t$ step.} 
	\label{alg-Pseudocode}  
	\begin{algorithmic}[1] 
		\Require  Image batch \emph{I}$_b$, $M^{t-1}$, $M^t$.
		
		\Ensure  
		Asymmetric region-wise contrastive loss $\mathcal{L}^{ARCL}$.
		\State $\mathcal{F}_{I_b}^{t-1}, pred_{I_b}^{t-1} \leftarrow M_{t-1}$(\emph{I$_b$}), $\mathcal{F}_{I_b}^{t}, pred_{I_b}^{t} \leftarrow M_{t}$(\emph{I}) \{forward pass\}
		\State $\mathcal{S}_{I_b}^{t-1}$=$argmax(pred_{I_b}^{t-1}, dim=1)$, \\  $\mathcal{S}_{I_b}^{t}$=$argmax(pred_{I_b}^{t}, dim=1)$ 
		\State $\mathcal{M} = torch.zeros(pred_{I_b}^{t-1}.shape)$ 
		\State \textbf{for} $c^i$ \textbf{in} $C^{1:t}$:  
		\State \qquad $\mathcal{M}_i^{t-1}[\mathcal{S}_{I_b}^{t-1}==c_i] = 1$ 
		\State \qquad $\mathcal{M}_i^t[\mathcal{S}_{I_b}^{t}==c_i] = 1$ 
		\State \qquad $\mathcal{M}_k^t[\mathcal{S}_{I_b}^{t-1}==c_k] = 1$ 
		\State \qquad \# filter region-wise embedding
		\State \qquad $\mathcal{E}_{a_i}^{t-1} =  \mathcal{M}_i^{t-1} \odot \mathcal{F}_{I_b}^{t-1}$ \{forward pass\} 
		\State \qquad $\mathcal{E}_{p_i}^{t} = \mathcal{M}_i^t \odot \mathcal{F}_{I_b}^{t}$ \{forward pass\}
		\State \qquad $\mathcal{E}_{n_i}^{t} = \mathcal{M}_k^t \odot \mathcal{F}_{I_b}^{t}$ \{forward pass\}
		\State \qquad ($\hat{\mathcal{E}}_{a_i}^{t-1}$, $\hat{\mathcal{E}}_{p_i}^{t}$, $\hat{\mathcal{E}}_{n_i}^{t}$) = ($\mathcal{E}_{a_i}^{t-1}$,  $\mathcal{E}_{p_i}^{t}$, $\mathcal{E}_{n_i}^{t}$).reshape(b, -1)
		\State \qquad  assert len($\hat{\mathcal{E}}_{a_i}^{t-1}$)==len($\hat{\mathcal{E}}_{p_i}^{t}$)==len($\hat{\mathcal{E}}_{n_i}^{t}$)
		\State \qquad $loss += \mathcal{L}_{ARCL}(\hat{\mathcal{E}}_{a_i}^{t-1}, \hat{\mathcal{E}}_{p_i}^{t},\hat{\mathcal{E}}_{n_i}^{t})$
		\State \textbf{return} $loss$
	\end{algorithmic}  
\end{algorithm}  
Particularly, the asymmetry manifests itself in two ways. One is the inequality of class number at the current step and previous step, the other is the embedding pairing manner for contrastive learning. Specifically, we select \textit{anchor} embedding from $M^{t-1}$ within old classes to reduce prediction error, while \textit{positive} and \textit{negative} embeddings from $M^t$ to optimize the current model. By filtering pixel embeddings belonging to the same class to a region, we achieve region-wise contrastive learning in those asymmetric latent spaces. Here we extend TripletMarginLoss~\cite{Hermans2017InDO} that considers intra-class compactness and inter-class dispersion with a margin between positive pairs and negative pairs. We construct positive pairs as $(\hat{\mathcal{E}}_{a_i}^{t-1}, \hat{\mathcal{E}}_{p_i}^{t})$ and negative pairs as $(\hat{\mathcal{E}}_{a_i}^{t}, \hat{\mathcal{E}}_{n_i}^{t})$. The optimization goal is to minimize the distance within positive pairs meanwhile maximize the distance within negative pairs. Thus the optimization goal is defined as: 
\begin{equation}
	\begin{aligned}
		&\mathcal{L}^{ARCL} = \\ &\frac{1}{N_c} \sum_{i}^{N_c} max\left(d(\hat{\mathcal{E}}_{a_i}^{t-1}, \hat{\mathcal{E}}_{p_i}^{t}) - d(\hat{\mathcal{E}}_{a_i}^{t}, \hat{\mathcal{E}}_{n_i}^{t}) + m, 0 \right) 
	\end{aligned}
\end{equation}
where $\emph{N$_c$}$ is the class numbers of the dataset. $d(\cdot)$ is a distance measurement function and $m$ is a constant.  We use L2-norm to construct $d(\cdot)$ in our implementation.
The pseudocode of the proposed ARCL is shown in Algorithm~\ref{alg-Pseudocode}.

\subsection{Dynamic Class-specific Pseudo Labelling}
\label{Sec-DCPL}
We present a Dynamic Class-specific Pseudo Labelling (DCPL) strategy.  The pioneer~\cite{PLOP} proposes a class-specific threshold but ignores the influence of semantic drift, in which case the degraded performance may reduce the confidence of pseudo labels, even causing negative optimization. To avoid overfitting incorrect pseudo labels, a class-specific threshold of every pixel is designed to preserve high-confidence pseudo labels for supervision.  

At $t$ step, the output segmentation map is $\mathcal{S}_{I}^{t} \in \mathbb{R}^{W\times H}$. We calculate the prediction score range $[u_c^l, u_c^h]$ for each class $c\in C^{0:t-1}$ in each training batch $I_b$ by $u_c^l=min(p_i(c))$ and $u_c^h=max(p_i(c))$, where $p_i(c)$=$soft\mathop{max}\limits_{c}(\mathcal{F}_{i}^{t})_{i\in I_b, c\in C^{0:t-1}}$ indicates the probability for class $c$ on pixel $i$.  Specifically, at $t$ step, the proposed class-specific threshold  $\tau_c^t$ is defined as:
\begin{equation}
	\begin{aligned}
		&  \tau_c^t= \left\{ 
		\begin{aligned}
				& u_c^l,  \quad &\rm{if} \ \bar{u}_c / \triangle \geq  \sigma \ and \ u_c^l \geq \epsilon \\ 
				& max(\Gamma, u_c^l) \quad &\rm{if} \ \bar{u}_c / \triangle <  \sigma \ and \ u_c^l \geq \epsilon  \\
				&\Gamma,  &\rm{otherwise} \quad \quad \quad\quad\quad\quad \  \\
			\end{aligned}
		\right. \\
		& \triangle = |u_c^h-u_c^l|, c\in C^{0:t-1}, \bar{u}_c = \frac{1}{n_c} \begin{matrix}\sum_{i\in I} p_i(c) \end{matrix}
		\end{aligned}
\end{equation}
where $p_i(c)$ indicates the score of $p_i$ at \textit{c}-th class. $\sigma$ measures the fluctuation of the model prediction on a class. Specifically, $ \bar{u}_c/\triangle \geq \sigma$ indicates high-confident predictions with small score fluctuation. $max(\Gamma, u_c^l)$ indicates preserving the most reliable predictions while $\bar{u}_c / \triangle <  \sigma$ means the unstable score fluctuation on class $c$. We simply set $\sigma=4$ based on experience. $\Gamma$ is a fixed threshold to avoid $\tau_c^t$ being too small. $\epsilon$ is the minimum confidence threshold and we set $\epsilon=0.5$ following the traditional protocol. $\sigma$ measures the score range of the prediction results from $M^{t-1}$. $n_c$ is the pixel number being predicted as class $c$. In this way, we try to balance preserving highly-confident pseudo labels with preserving as many pseudo labels as possible.
Thus the pseudo-label $\mathfrak{L}_{pl, c}^{t}$ for class $c$ can be generated from $M^{t-1}$:
\begin{equation}
    \begin{aligned}
    &\mathfrak{L}_{pl, c}^{t} = \mathop{\mathfrak{I}}\limits_{i}^{W\times H} \left[ \left(arg\mathop{max}\limits_{k} p_i(k)=c\right) \wedge  \left(p_i(c) > \tau_{c}^t \right) \right], \\ 
    &c, k\in C^{0:t-1}, i \in \mathcal{S}_{I}^{t}
    \end{aligned}
\end{equation}
where $\mathfrak{I}$ represents traversing pixels on $\mathcal{S}_{I}^{t}$. In this case, the high threshold ensures high-confident pseudo labels for \emph{easy} classes. While the pseudo labels for \emph{hard} classes have the least errors.  Since only ground truth of $C^t$ is accessed at $t$ step training, we concatenate $\mathfrak{L}_{pl}^{t}$ and $\mathfrak{L}^{t}$ to supervise $M_t$:
\begin{equation}
	\mathfrak{L}_{s}^{t} =\sum_{c}^{C^{0:t-1}} \mathfrak{L}_{pl, c}^{t} \oplus \mathfrak{L}^t 
\end{equation}
where $\mathfrak{L}^{t}$ indicates the ground truth of $C^t$, $\oplus$ represents the label combination operation.  Under the supervision of  $\mathfrak{L}_{s}^{t}$, the segmentation loss for image $I^t$ at $t$ step training can be calculated by cross-entropy (CE) loss:
\begin{equation}
	\mathcal{L}^{Seg}(I^t, \mathfrak{L}_s^t) =  - \sum_{i\in I^t} y_{ik}log(p_{ik})
\end{equation}
where $y_{ik} \in  \mathfrak{L}_s^t$ and $p_{ik}$ represent the ground truth probability and the predicted probability of class $k$ on pixel $i$, respectively.
Aiming at addressing the CISS in an end-to-end way, we perform the above ARCL and DADA synchronously. The integrated objective is defined as: 
\begin{equation}
	\mathcal{L}^{total} = \mathcal{L}^{Seg} + \mathcal{L}^{DADA} + \mathcal{L}^{ARCL} 
\end{equation}

Particularly, at the initial step, only $\mathcal{L}^{Seg}$ is calculated, which is the same as the static training process. While at each incremental step, $\mathcal{L}^{total}$ is calculated for the whole network training. 

\section{Experiments}
\label{Sec-Experiment}
\subsection{Datasets and Protocols}
\noindent\textbf{Datasets}. 1) PASCAL VOC 2012~\cite{VOC2012} is a widely used benchmark for semantic segmentation. It consists of 10582 training images and 1449 images for validation with 20 semantic classes and an extra background class.  We evaluate our model on 15-5 (2 steps), 15-1 (6 steps), 5-3 (6 steps) and 10-1 (11 steps) settings. For example, 15-1 means initially learning 15 classes and learning additional 5 classes at another step. 15-1 indicates initially learning 15 classes and then learning the additional one class at each step for a total of another 5 steps. 
2) ADE20K~\cite{ADE} is a large-scale semantic segmentation dataset containing 150 classes that cover indoor and outdoor scenes. The dataset is split into 20210 training images and 2000 validation images. Compared to Pascal VOC 2012, ADE20K covers a wider variety of classes in natural scenes with more instances and categories. We evaluate our model on 100-50 (2 steps), 100-10 (6 steps), 50-50 (3 steps) and 100-5 (11 steps) settings. 
3) ISPRS~\cite{2013ISPRS} consists of two airborne image datasets including Postdam and Vaihingen. We conduct our experiments on Postdam.  It contains 38 images with the size of 6000$\times$6000 pixels. The resolution of each image is 5 cm. As the source organizer suggested, we take 24 images for training and 14 images for validation. For training convenience, we partition each image into 600$\times$600 patches in sequential order.  Semantic content in Postdam has been classified manually into six land cover classes, namely: \textit{imprevious surfaces}, \textit{building}, \textit{low vegetation}, \textit{tree}, \textit{car} and \textit{clutter}. Following previous works~\cite{Li2022ClassIncrementalLN, Shan2022ClassIncrementalLF}, we choose to ignore class \textit{clutter} in the training and validation process since it only accounts for very few pixel quantities and its unclear semantic scope.  We evaluate our model on 4-1 (2 steps), 2-3 (2 steps), 2-2-1 (3 steps) and 2-1 (4 steps) settings.

\noindent\textbf{Metrics}. We compute Pascal VOC mean intersection-over-union (mIoU)~\cite{VOC2012} for evaluation: 
\begin{equation}
	mIoU = \frac{TP}{TP+FP+FN}
\end{equation}
where TP, FP and FN are the numbers of true positive, false positive and false negative pixels, respectively.

Specifically, we compute mIoU after the last step T for the initial classes $C^{0: init}$, for the incremented classes $C^{init+1:T}$, and for all classes $C^{0:T}$ (all). 

\noindent \textbf{Protocols}. Following~\cite{MiB, PLOP}, there are two different incremental settings: \emph{disjoint} and \emph{overlapped}. In both settings, only the current classes $C^t$ are labelled and an extra background (bg) class $C^{bg}$. In the former, images at $t$ step only contain $C^{0:t-1} \cup C^t \cup C^{bg}$. While the latter contains $C^{0:t-1} \cup C^{t} \cup C^{t+1: T} \cup C^{bg}$, which is more realistic and challenging. In this study, we focus on \emph{overlapped} setting in our experiments. We report two baselines for reference, i.e., \emph{fine tuning} on $C^{t}$,  and training on all classes \emph{offline}. The former is the lower bound and the latter can be regarded as the upper bound in CISS tasks. 

\subsection{Implementation Details}
We use DeepLabv3~\cite{DeepLabv3} architecture with a ResNet-101~\cite{ResNet} backbone pretrained on ImageNet~\cite{ImageNet} as semantic segmentation baseline, as same as~\cite{PLOP, UCD}. For all experiments, the initial learning rate is 0.01 and decayed by a \emph{poly} learning rate policy. The training batch size is set to 12 in all CISS settings. At each step,  we use SGD~\cite{SGD} as the optimizer with a momentum of 0.9 and a weight decay of 5$\times$10$^{-4}$. We train the model with 30 (Pascal VOC 2012, ISPRS) and 50 (ADE20K) epochs for each incremental step, respectively. The input image is resized to 513$\times$513. Standard data augmentation methods are applied during training, such as random scaling from 0.5 to 2.0 and random flipping. For hyper-parameters, we set $\Gamma=0.7$, $\gamma=0.9$ and $\lambda=2.0$ according to the analysis in Sec.~\ref{abla-parameters}. $\alpha=1.0$. Implementation of the proposed network is based on Pytorch 1.8 with CUDA 11.6 and all experiments are conducted on a workstation with 4 NVIDIA 3090 GPUs.  The code will be available at our formal published version.

\subsection{Quantitative Evaluation}
\noindent\textbf{Pascal VOC 2012}. We compare our method with current state-of-the-art data-free approaches including~\cite{EWC, LWF, ILT, 2021Knowledge, MiB, PLOP, SDR, UCD, REMINDER, RCIL} and data-replay-based methods~\cite{RECALL, SSUL}. Table~\ref{table-VOC2012} shows quantitative experiments on VOC 15-5, 15-1, 5-3 and 10-1 settings. In multi-step CISS tasks like 15-1, 5-3 and10-1, our method is superior to the closest contender~\cite{RCIL}. For example, mIoU on all $C^{0:20}$ of the proposed IDEC is 67.32\% in 15-1 case, of which the performance of $C^{0:15}$ achieves a large improvement than the competitors, proving the effectiveness of our distillation mechanism. And our performance of the incremental $C^{16:20}$ is superior to~\cite{PLOP} by a significant margin, demonstrating the adaptive capacity on new classes. Since DADA and ARCL share a mutually reinforced relationship as introduced in Sec.~\ref{Sec-ARCL}, the experimental results also prove the cooperativity between the DADA and ARCL. 
It is worth mentioning that the proposed model achieves a performance close to the previous replay-based approaches like~\cite{RECALL}, validating the effectiveness of the proposed strategy.

\begin{figure}
	\centering
	\includegraphics[scale=0.53]{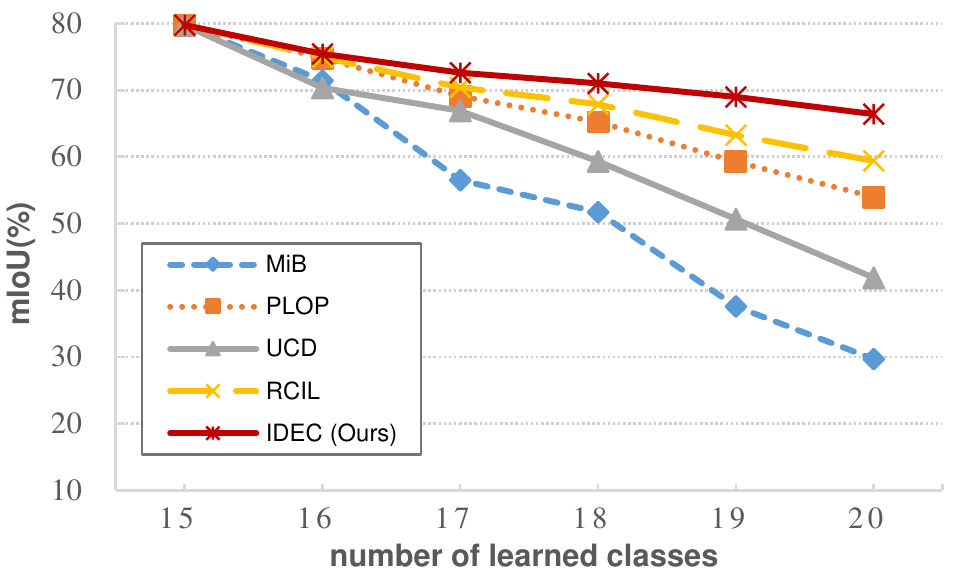}
	\caption{The mIoU (\%) evolution against the number of learned classes in VOC 15-1 task. }
	\label{fig-mIoU_curve}
\end{figure}

\begin{figure*}[htbp]
	\centering
	\includegraphics[scale=0.67]{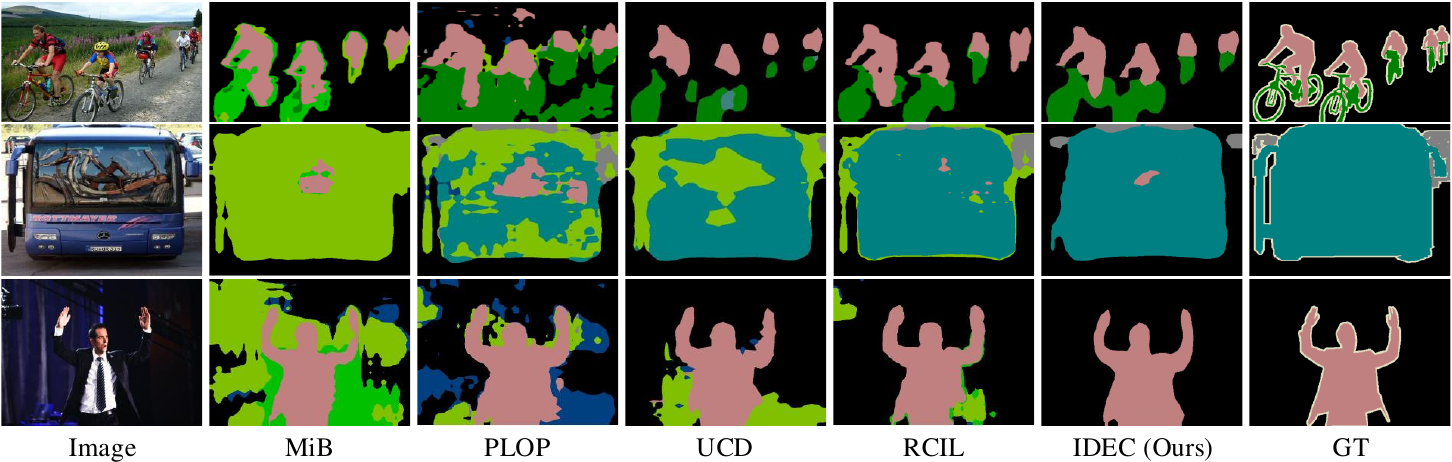}
	\caption{Qualitative results between various approaches in VOC 15-1 task.}
	\label{fig-VC}
\end{figure*}

In Fig.~\ref{fig-mIoU_curve}, we evaluate mIoU against the number of learned classes on VOC 15-1. For example, MiB~\cite{MiB} degenerates rapidly at the latter steps. Current state-of-the-art~\cite{RCIL} is also confronted with severe performance degradation. The proposed IDEC maintains the highest mIoU among the contrast models after all incremental steps, indicating the strong ability against forgetting and adaptation to new classes. Fig.~\ref{fig-VC} shows the visualization comparison on VOC 15-1 task. The proposed method maintains the region integrity of different classes after all IL steps.

\noindent\textbf{ADE20K}. Table~\ref{table-ADE20K} shows quantitative experiments on ADE 100-50, 100-10, 50-50 and 100-5 tasks. Due to the large number of classes, it is more challenging thus the upper bound mIoU is only 38.9\%, indicating that there is severe pixel misclassification. The representative PLOP~\cite{PLOP} shows robustness in 100-50 and 50-50 with fewer IL steps. In contrast, the proposed IDEC shows robust learning ability in more challenging multi-step CISS tasks like 100-10 and 100-5. For example, in 100-5,  IDEC shows strong ability against catastrophic forgetting as the mIoU of $C^{1:100}$ is the highest among all contenders and finally achieves 31.00\% mIoU on all classes. 
In addition, compared with the first-tier replay-based method~\cite{SSUL}, the proposed method also shows competitive performance in anti-forgetting ability and new-class adaptation.

\begin{figure*}[htbp]
	\centering
	\includegraphics[scale=1.10]{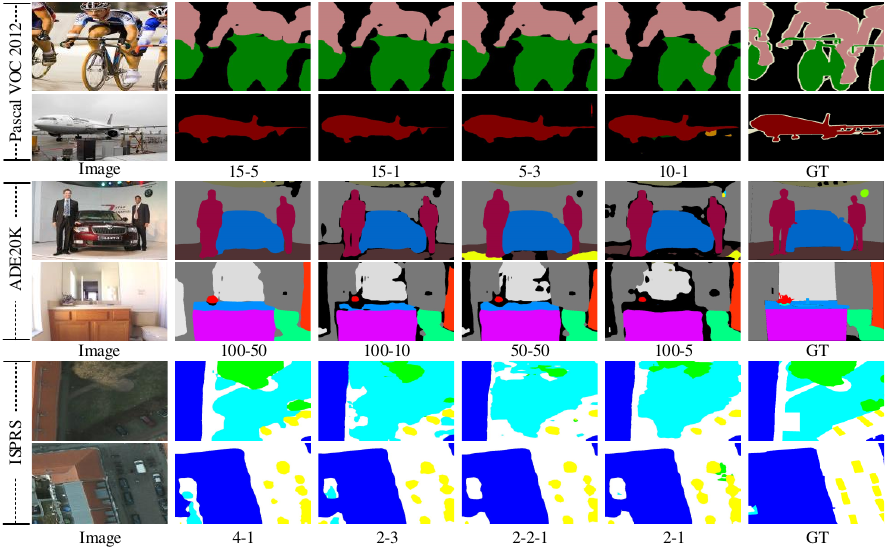}
	\caption{Visualization results of our method on Pascal VOC 2012, ADE20K and ISPRS with various incremental settings.}
	\label{fig-VT}
\end{figure*}

\noindent\textbf{ISPRS}. We extend our method to remote-sensing scenes. Space-based in-orbit remote sensing is a suitable field for model deployment with incremental learning. Compared to natural scenes, remote sensing image contains more terrain context including complex noise interference, occlusion issue and obscure region boundaries, etc. Table~\ref{table-ISPRS} displays the comparison of~\cite{EWC, ILT, MiB, PLOP} and our method. The proposed IDEC achieves the highest mIoU on ISPRS 4-1, 2-3, 2-2-1 and 2-1 tasks. For example, with respect to the most challenging task 2-1, our method is superior to MiB~\cite{MiB} with 7.06\% mIoU. on 2-3, 2-2-1 and 2-1 tasks, IDEC remains a competitive anti-forgetting performance on the initial two classes \textit{imprevious surfaces} and \textit{building}, validating the anti-forgetting ability.
Fig.~\ref{fig-VT} displays representative visualization results on multiple CISS tasks on all three datasets. During the IL steps, IDEC maintains the region integrity of different classes after all IL steps and also shows advantages in reducing pixel misclassification.
 
\begin{table*}[htbp]
	\caption{Quantitative comparison on Pascal VOC 2012 in mIoU (\%).  Class 0 indicates the unlabelled class. The first and second best results are highlighted in \textcolor{red}{red} and \textcolor{blue}{blue}, respectively. * indicates the results are from~\cite{PLOP}. $\ddagger$ means we re-implement the method. }
	\centering
	\footnotesize
	\setlength{\tabcolsep}{1mm}{
		{\begin{tabular*}{0.95\textwidth}{@{\extracolsep{\fill}}ll|ccc|ccc|ccc|ccc@{}}
				\toprule[0.5mm]
				&\multirow{2}*{Method} &
				\multicolumn{3}{c}{15-5 (2 steps)} & \multicolumn{3}{c}{15-1 (6 steps)} & \multicolumn{3}{c}{5-3 (6 steps)} & \multicolumn{3}{c}{10-1 (11 steps)} \\
				
				&& 0-15 & 16-20 & all & 0-15 & 16-20 & all & 0-5 & 6-20 & all & 0-10 & 11-20 & all\\
				\midrule
				\multirow{13}*{Data-free} &\emph{fine tuning}&2.10&33.10&9.80&0.20&1.80&0.60&0.50&10.40&7.60&6.30&2.80&4.70\\
				&EWC*~\cite{EWC} &24.30 &35.50 &27.10 &0.30 &4.30 &1.30 &-&-&-&- &- &-\\
				&LwF-MC*~\cite{LWF} &58.10&35.00&52.30&6.40&8.40&6.90&20.91&36.67&24.66&4.65&5.90&4.95\\
				&ILT*~\cite{ILT} &66.30 &40.60&59.90&4.90&7.80&5.70&22.51&31.66&29.04&7.15&3.67&5.50\\
				&Umberto \emph{et al.}$^\ddagger$~\cite{2021Knowledge}&67.20&38.42&60.35&8.75&7.99&8.56&26.15&37.84&34.50&6.52&5.16&5.87\\
				&MiB*~\cite{MiB}&76.37&49.97&70.08&34.22&13.50&29.29&57.10&\textcolor{blue}{42.56}&46.71&12.25&13.09&12.65\\
				&PLOP*~\cite{PLOP}&75.73&51.71&70.09&65.12&21.11&54.64&17.48&19.16&18.68&44.03&15.51&30.45 \\
				&SDR~\cite{SDR} &75.40&\textcolor{blue}{52.60}&69.90&44.70&21.80&39.20&-&-&-&32.40&17.10&25.10 \\
				&UCD$^\ddagger$~\cite{UCD}&77.50&\textcolor{red}{\textbf{53.10}}&71.30&49.00&19.50&41.90&31.35&23.43&25.69&38.73&22.52&31.01\\
				&UCD+PLOP~\cite{UCD}&75.00&51.80&69.20&66.30&21.60&55.10&-&-&-&42.30&\textcolor{blue}{28.30}&\textcolor{blue}{35.30}\\
				&REMINDER~\cite{REMINDER} &76.11&50.74&70.07&68.30&\textcolor{blue}{27.23}&58.52&-&-&-&-&-&- \\
				&RCIL$^\ddagger$~\cite{RCIL}&\textcolor{red}{\textbf{78.80}}&52.00&\textcolor{red}{\textbf{72.40}}&\textcolor{blue}{70.60}&23.70&\textcolor{blue}{59.40}&\textcolor{blue}{65.30}&41.49&\textcolor{blue}{50.27}&\textcolor{blue}{55.40}&15.10&34.30\\
				\cmidrule{2-14} 
				&\textbf{IDEC} (Ours) 
				&\textcolor{blue}{78.01}&51.84&\textcolor{blue}{71.78}&\textcolor{red}{\textbf{76.96}}&\textcolor{red}{\textbf{36.48}}&\textcolor{red}{\textbf{67.32}}&\textcolor{red}{\textbf{67.05}}&\textcolor{red}{\textbf{48.98}}&\textcolor{red}{\textbf{54.14}}&\textcolor{red}{\textbf{70.74}}&\textcolor{red}{\textbf{46.30}}&\textcolor{red}{\textbf{59.10}}\\
				\midrule
				\multirow{3}*{Data-replay} &RECALL-GAN~\cite{RECALL}&66.60&50.90&64.00&65.70&47.80&62.70&-&-&-&59.50&46.70&54.80\\
				& RECALL-Web~\cite{RECALL}&67.70&54.30&65.60&67.80&50.90&64.80&-&-&-&65.00&53.70&60.70\\
				&SSUL-M~\cite{SSUL}&79.53&52.87&73.19&78.92&43.86&70.58&72.97&49.02&55.85&74.79&48.87&65.45\\
				\midrule
				&\emph{offline} &79.77&72.35&77.43&79.77&72.35&77.43&76.91&77.63&77.43&78.41&76.35&77.43\\
				\bottomrule[0.5mm]
		\end{tabular*}}{}}	
	\label{table-VOC2012}
\end{table*}

\begin{table*}[htbp]
	\centering
	\footnotesize
	\caption{Quantitative comparison on ADE20K in mIoU (\%).  * indicates the results are from~\cite{PLOP}. The first and second best results are highlighted in \textcolor{red}{red} and \textcolor{blue}{blue}, respectively. $\ddagger$ means we re-implement the method.}
	\setlength{\tabcolsep}{1mm}{
		{\begin{tabular*}{0.95\textwidth}{@{\extracolsep{\fill}}ll|ccc|ccc|ccc|ccc@{}}
				\toprule[0.5mm]
				&\multirow{2}*{Method} &
				\multicolumn{3}{c}{100-50 (2 steps)}   & \multicolumn{3}{c}{100-10 (6 steps)} & \multicolumn{3}{c}{50-50 (3 steps)} & \multicolumn{3}{c}{100-5 (11 steps)}\\
				&& 1-100 & 101-150 & all & 1-100 & 101-150 & all & 1-50 & 51-150 & all  & 1-100 & 101-150 & all\\
				\midrule
				\multirow{9}*{Data-free}&\emph{fine tuning}&0.00&11.22&3.74&0.00&2.08&0.69&0.00&3.60&2.40&0.00&0.07&0.02\\
				&ILT*~\cite{ILT} &18.29&14.40&17.00&0.11&3.06&1.09 &3.53&12.85&9.70&0.08&1.31&0.49\\
				&Umberto \emph{et al.}$^\ddagger$~\cite{2021Knowledge}&22.79&13.81&19.80&3.28&5.44&4.00&8.89&14.40&12.56&1.02&1.25&1.10\\
				&MiB*~\cite{MiB}&40.52&17.17&32.79&38.21&11.12&29.24&45.57&21.01&29.31&36.01&5.66&25.96\\
				&PLOP*~\cite{PLOP}&41.87&14.89&32.94&\textcolor{blue}{40.48}&13.61&31.59&\textcolor{red}{\textbf{48.83}}&20.99&30.40&\textcolor{blue}{39.11}&7.81&28.75\\
				&UCD+PLOP~\cite{UCD}&\textcolor{blue}{42.12}&15.84&33.31&\textcolor{red}{\textbf{40.80}}&15.23&32.29&47.12&24.12&31.79&-&-&-\\
				&REMINDER~\cite{REMINDER} &41.55&\textcolor{red}{19.16}&\textcolor{blue}{34.14}&38.96&\textcolor{red}{\textbf{21.28}}&\textcolor{red}{\textbf{33.11}}&47.11&20.35&29.39&36.06&\textcolor{red}{\textbf{16.38}}&29.54 \\
				&RCIL~\cite{RCIL}&\textcolor{red}{\textbf{42.30}}&\textcolor{blue}{\textbf{18.80}}&\textcolor{red}{\textbf{34.50}}&39.30&17.60&32.10&\textcolor{blue}{48.30}&\textcolor{blue}{25.00}&\textcolor{blue}{32.50}&38.50&11.50&\textcolor{blue}{29.60}\\
				\cmidrule{2-14} 
				&\textbf{IDEC} (Ours) 
				&42.01&18.22&34.08&40.25&\textcolor{blue}{17.62}&\textcolor{blue}{32.71}&47.42&\textcolor{red}{\textbf{25.96}}&\textcolor{red}{\textbf{33.11}}&\textcolor{red}{\textbf{39.23}}&\textcolor{blue}{14.55}&\textcolor{red}{\textbf{31.00}}\\
				\midrule
				\multirow{1}*{Data-replay}&SSUL-M~\cite{SSUL}&42.20&13.95&32.80&42.17&16.03&33.89&49.55&25.89&33.78&42.53&15.85&34.00\\
				\midrule
				&\emph{offline} &44.30&28.20&38.90&44.30&28.20&38.90&50.90&32.90&38.90&44.30&28.20&38.90\\
				\bottomrule[0.5mm]
		\end{tabular*}}{}}	
	\label{table-ADE20K}
\end{table*}

\begin{table*}[htbp]
	\centering
	\footnotesize
	\caption{Quantitative comparison on ISPRS in mIoU (\%).  The first and second best results are highlighted in \textcolor{red}{red} and \textcolor{blue}{blue}, respectively. $\ddagger$ means we re-implement the method by the code provided in~\cite{PLOP}.}
	\setlength{\tabcolsep}{1mm}{
		{\begin{tabular*}{0.95\textwidth}{@{\extracolsep{\fill}}l|ccc|ccc|ccc|ccc@{}}
				\toprule[0.5mm]
				\multirow{2}*{Method} &
				\multicolumn{3}{c}{4-1 (2 steps)}   & \multicolumn{3}{c}{2-3 (2 steps)} & \multicolumn{3}{c}{2-2-1 (3 steps)} & \multicolumn{3}{c}{2-1 (4 steps)}\\
				& 1-4 & 5 & all & 1-2 & 3-5 & all & 1-2 & 3-5 & all  & 1-2 & 3-5 & all\\
				\midrule
				\emph{fine tuning}&2.75&30.28&8.26&5.60&19.22&13.77&0.53&7.90&4.95&0.02&1.05&0.64\\
				EWC$^\ddagger$~\cite{EWC} &27.01&31.25&27.86&39.59&25.94&31.40&17.88&10.46&13.43&8.97&2.35&5.00\\
				ILT$^\ddagger$~\cite{ILT} & 42.37&30.92&40.08&41.29&25.83&32.01&35.28&19.77&25.97&20.92&8.44&13.43\\
				MiB$^\ddagger$~\cite{MiB}&\textcolor{blue}{67.11}&\textcolor{blue}{38.90}&\textcolor{blue}{61.47}&\textcolor{blue}{76.49}&\textcolor{blue}{44.45}&\textcolor{blue}{57.27}&\textcolor{blue}{67.83}&39.07&50.57&\textcolor{blue}{65.28}&\textcolor{blue}{41.46}&\textcolor{blue}{50.99}\\
				PLOP$^\ddagger$~\cite{PLOP}&65.78&36.12&59.85&70.44&43.80&54.46&66.45&\textcolor{blue}{41.68}&\textcolor{blue}{51.59}&64.41&40.89&50.30 \\
				\midrule 
				\textbf{IDEC} (Ours)
				&\textcolor{red}{\textbf{74.05}}&\textcolor{red}{\textbf{50.77}}&\textcolor{red}{\textbf{69.39}}&\textcolor{red}{\textbf{78.91}}&\textcolor{red}{\textbf{49.08}}&\textcolor{red}{\textbf{61.01}}&\textcolor{red}{\textbf{75.45}}&\textcolor{red}{\textbf{47.54}}&\textcolor{red}{\textbf{58.70}}&\textcolor{red}{\textbf{76.25}}&\textcolor{red}{\textbf{45.89}}&\textcolor{red}{\textbf{58.03}}\\
				\midrule
				\emph{offline} &75.23&63.97&72.98&83.31&66.09&72.98&83.31&66.09&72.98&83.31&66.09&72.98\\
				\bottomrule[0.5mm]
		\end{tabular*}}{}}	
	\label{table-ISPRS}
\end{table*}

\begin{figure*}[htbp]
	\centering
	\includegraphics[scale=0.64]{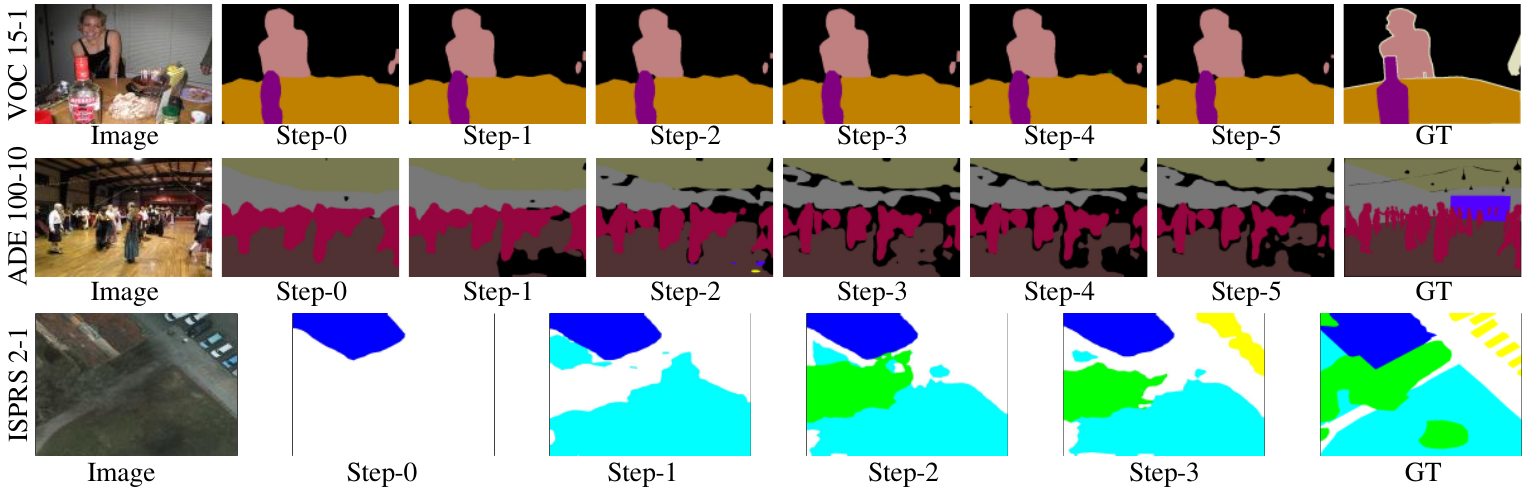}
	\caption{Step-wise visualization results of the proposed method in terms of  VOC 15-1, ADE 100-10 and ISPRS 2-1. }
	\label{fig-VS}
\end{figure*}

\subsection{Ablation Study}
\subsubsection{Module Contribution}
\label{Sec-MC}
\textbf{Distillation Mechanism}. As introduced in Sec.~\ref{Sec-DADA}, we propose a distillation mechanism considering both intermediate layers and output logits. In Table~\ref{table-Aba-DM} we study the performance benefits of module contributions covering VOC 15-1, ADE 100-10 and ISPRS 2-1.  For the distillation strategy, we present two settings, 1) OL-D after IL-D; 2) IL-D after OL-D, to reveal the contribution of KD in CISS tasks. Since the pseudo labels are generated by the old model to boost the latter-step learning, the IoU performance would reveal the impact of orders of IL-D and OL-D. In all three CISS tasks, firstly adopting IL-D then OL-D achieves higher IoU than the other setting. In our opinion, this validates the benefits of distilling low- and high-level features synchronously to CISS. 
Taking VOC 15-1 as an example, without distillation but only performing CE on $C^t$, the performance is only 0.60\% mIoU. With intermediate layer distillation (IL-D). and pseudo-labelling without DCPL (but with a fixed value $\tau_c=0.7$), the mIoU increases to 61.44\%. It also proves that ALW can boost the layer-wise distillation with an extra 0.76\% mIoU improvement. And with additional output logits distillation (OL-D), it gains another 3.60\% improvement.  Thus in the following experiments, we adopt the OL-D after IL-D setting with the ALW guidance.

\begin{table}[h]
	\centering
	\caption{The contribution of proposed modules. The results come from Pascal VOC 2012 15-1, ADE20K 100-10 and ISPRS 2-1. When not using DCPL, the pseudo labels are generated by a fixed threshold $\tau_c=0.7$.}
	\setlength{\tabcolsep}{0.35mm}{
		\begin{tabular}{l|ccc|ccc|ccc}
			\toprule[0.4mm]
			\multirow{2}*{Method} &\multicolumn{3}{c}{VOC 15-1} & \multicolumn{3}{c}{ADE 100-10} & \multicolumn{3}{c}{ISPRS 2-1} \\
			& 0-15 & 16-20 & all & 1-100 & 101-150 & all &1-2&3-5&all \\
			\midrule
			\emph{fine tuning} &0.20&1.80&0.60&0.00&2.08&0.69 &0.02&1.05&0.64\\
			\midrule
            +IL-D w/o ALW&71.86&28.11&61.44&36.66&15.47&29.60 &69.79&37.68&50.52\\
            +IL-D w/ ALW&72.73&28.49&62.20&37.33&15.35&30.00 &69.55&38.26&50.78\\
            +OL-D &74.96&36.48&65.80&38.12 &17.06&31.10&73.59&42.32&54.83 \\
			\midrule
			+OL-D&71.87&26.65&61.10&34.98&15.21&28.39&69.48&38.35&50.80\\
			+IL-D w/o ALW&73.12&35.51&64.17&37.10&16.73&30.31&71.25&41.55&53.43\\
			+IL-D w/ ALW&73.65&35.68&64.61&38.05&16.73&30.94&73.28&41.89&54.45\\
			\midrule
			+ARCL &75.21&\textbf{38.00}&66.35&40.12&17.44&32.56 &75.45&45.54&57.50\\
			\midrule
			+DCPL&\textbf{76.96}&36.48&\textbf{67.32}&\textbf{40.25}&\textbf{17.62}&\textbf{32.71}&\textbf{76.25}&\textbf{45.89}&\textbf{58.03}\\
			\bottomrule[0.4mm]
	\end{tabular}}
	\label{table-Aba-DM}
\end{table}
\begin{table}[h]
	\centering
	\caption{The mIoU comparison on pseudo-labelling strategy by applying no threshold (none), fixed threshold (we set $\tau_c=0.7$), PLOP~\cite{PLOP} pseudo-labelling strategy and the proposed DCPL.}
	\setlength{\tabcolsep}{1.2mm}{
		\begin{tabular}{l|cccc|c}
			\toprule[0.4mm]
			Task&none&fixed&PLOP$^{\ddagger}$&DCPL&mIoU \\
			\midrule
			\multirow{4}{*}{\makecell[c]{VOC\\15-1}}&\checkmark&&&&61.87 \\
			&&\checkmark&&&66.35\\
			&&&\checkmark&&65.44 \\
			&&&&\checkmark&\textbf{67.32}\\
			\midrule
			\multirow{4}{*}{\makecell[c]{ADE\\100-10}}&\checkmark&&&&29.17\\
			&&\checkmark&&&32.56\\
			&&&\checkmark&&31.71\\
			&&&&\checkmark&\textbf{32.71}\\
			\midrule
			\multirow{4}{*}{\makecell[c]{ISPRS\\2-1}}&\checkmark&&&&54.68 \\
			&&\checkmark&&&57.50 \\
			&&&\checkmark&&55.41\\
			&&&&\checkmark&\textbf{58.03} \\
			\bottomrule[0.4mm]
		\end{tabular}
	}
	\label{table-Aba-DCPL}
\end{table}

\noindent\textbf{Contrastive Learning}. Compared to the pioneers~\cite{SDR, UCD}, the proposed ARCL constructs contrastive learning through high-confidence embeddings to avoid the misleading caused by model degeneration. Based on the proposed KD strategies, we further evaluate the proposed contrastive learning efficiency in Table.~\ref{table-Aba-DM}. With respect to VOC 15-1 task, ARCL brings 0.55\% mIoU improvement in all classes. In detail, it achieves 0.25\% and 1.52\% mIoU advancement on $C^{0:15}$ and the incremental classes $C^{16:20}$, respectively. For ADE 100-10, ARCL achieves 1.46\% mIoU improvement based on the KD strategy. As for ISPRS 2-1 task,  the proposed ARCL also proves its robustness by bringing 2.07\% mIoU advancement. From the analysis above, ARCL shows a robust ability on alleviating forgetting and classifier bias during IL steps.

\noindent\textbf{Pseudo-labelling}. Standing on the shoulders of the above two modules, we explore the contribution of the proposed DCPL. On the one hand, DCPL brings positive earnings in all classes on all three tasks. As seen in Table~\ref{table-Aba-DM}, compared to a fixed threshold, i.e., $\tau_c=0.7$, DCPL brings 0.97\%, 0.15\% and 0.53\% mIoU improvement on VOC 15-1, ADE 100-10 and ISPRS 2-1, respectively. However, there is an exception that in VOC 15-1, it brings 1.75\% IoU improvement on $C^{0:15}$ but 1.52\% negative earnings on $C^{16:20}$. This reveals that in CISS, it is challenging to balance the anti-forgetting and new-class adaptation if the segmentation model achieves low performance, which is discussed in Sec.~\ref{Sec-Abla-AFNA}. On the other hand,  Table~\ref{table-Aba-DCPL} compares the efficiency of several pseudo-labelling strategies.  In comparison with the uncertainty-based pseudo-labelling strategy in~\cite{PLOP}, the proposed approach also shows competitive efficiency. In line with our expectations, the proposed dynamic class-specific threshold benefits the IL performance because of the balance of pixel misclassification reduction and pseudo-label supplement. 
\begin{figure*}[ht]
	\includegraphics[scale=0.33]{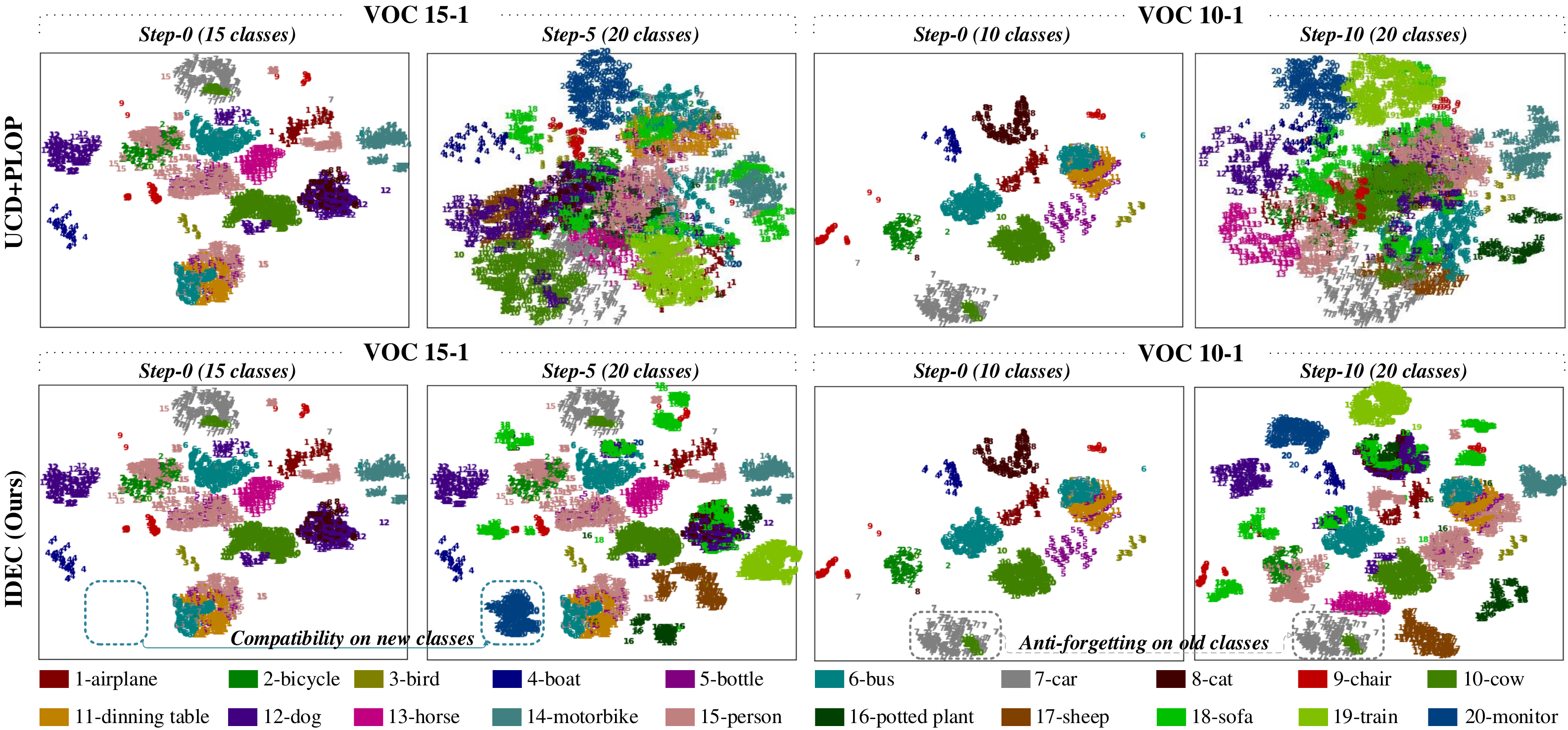}
	\caption{T-SNE~\cite{TSNE} visualizations on VOC 15-1 and 10-1, best viewed in colour. For visualization convenience, the background class is skipped. The number in this image represents the corresponding class.}
	\label{fig-tsne}
\end{figure*}

\subsubsection{Anti-forgetting and New-class Adaptation}
\label{Sec-Abla-AFNA}
As mentioned in Sec.~\ref{Sec-Introduction}, models after incremental steps face catastrophic forgetting. Thus the anti-forgetting ability in CISS is of great concern. We evaluate the mIoU for initially learned classes (those learned at Step-0) after each incremental step on VOC 15-1, ADE 100-10 and ISPRS 2-1, respectively. Taking VOC 15-1 as an example, we re-evaluated the mIoU of $C^{0:15}$ using the new model at each step in Table~\ref{table-Aba-antiforgetting}.  It is observed that the mIoU of  $C^{0:15}$ decreases step by step, which is in accordance with our expectations. While it also shows a strong anti-forgetting ability since there is only 2.81\% mIoU decline after the final step. In terms of ADE 100-10, which is more challenging due to the large number of classes, the mIoU of $C^{1:100}$ drops 2.52\% after the final step. IDEC also maintains the most capacity on the old classes. In ISPRS 2-1, the anti-forgetting is more challenging since the remote-sensing images normally contain more complex semantics and context association. At Step-0, the mIoU of  $C^{1:2}$ achieves 91.88\% and drops rapidly to 79.79\% at Step-1, but finally maintains 76.07\% at Step-3. We also compared the proposed method to the pioneer~\cite{PLOP},  and our method shows a more robust anti-forgetting performance on all three CISS tasks.
\begin{table}[h]
	\centering
	\caption{The mIoU (\%) evolution of initially learned classes with the increase of learning steps. We evaluate VOC 15-1 (6 steps), ADE 100-10 (6 steps) and ISPRS 2-1 (4 steps). $\ddagger$ indicates re-implementing the method. }
	\setlength{\tabcolsep}{1.2mm}{
		\begin{tabular}{l|cccccc}
			\toprule[0.4mm]
			\multirow{2}*{task} &\multicolumn{6}{c}{VOC 15-1 mIoU$_{C^{0:15}}$}  \\
			
			&Step-0 & Step-1 &Step-2  &Step-3 &Step-4 &Step-5\\
			\midrule
			PLOP$^\ddagger$~\cite{PLOP}&79.77&72.95&68.42&66.76&65.52&64.27\\
			IDEC &79.77&\textbf{79.32}&\textbf{78.21}&\textbf{77.60}&\textbf{77.05}&\textbf{76.96}\\
			\midrule
			\multirow{2}*{task} &\multicolumn{6}{c}{ADE 100-10 mIoU$_{C^{1:100}}$} \\
			&Step-0 & Step-1 &Step-2  &Step-3 &Step-4 &Step-5\\
			\midrule
			PLOP$^\ddagger$~\cite{PLOP}&42.77&41.38&40.71&40.52&\textbf{40.41}&40.13\\
			IDEC&42.77&\textbf{41.79}&\textbf{41.00}&\textbf{40.67}&40.40&\textbf{40.25}\\
			\midrule
			\multirow{2}*{task} &\multicolumn{6}{c}{ISPRS 2-1 mIoU$_{C^{1:2}}$} \\
			&Step-0 & Step-1 &Step-2  &Step-3 &-&-\\	
			\midrule
			PLOP$^\ddagger$~\cite{PLOP}&91.95&76.53&73.24&64.41&-&-\\
			IDEC&91.95&\textbf{79.36}&\textbf{77.89}&\textbf{76.25}&-&-\\
			\bottomrule[0.4mm]
	\end{tabular}}
	\label{table-Aba-antiforgetting}
\end{table}

Since we aim to stimulate the model to remember old classes and learn new classes concurrently, these two sides should boost each other. In dense prediction tasks, a pixel only can be assigned one label. It means with a favourable anti-forgetting ability, the model can achieve high-confident pseudo labels. Besides, since we tackle the catastrophic forgetting and semantic drift synergistically, a solid anti-forgetting ability is also conducive for the latter contrastive learning to improve the adaptation ability on new classes, which is validated in Sec.~\ref{Sec-MC}. To investigate the inner feature distribution after incremental steps, we use t-SNE~\cite{TSNE} to map the high-dimensional features to 2D space. We evaluate our method on VOC 15-1 and 10-1 at the initial step and the final step. 
As seen in Fig.~\ref{fig-tsne},  the position is anchored for each class. We demonstrate the advantage of the proposed method compared with UCD+PLOP~\cite{UCD} in inner feature distribution. On VOC 15-1, UCD+PLOP is invaded by catastrophic forgetting resulting in class confusion at the final step. IDEC achieves consistent classification ability on the initial $C^{0:15}$ at Step-0 and Step-5. While new classes $C^{16:20}$ also can be distinguished from old classes and be classified into new clusters. It intuitively proves IDEC can effectively boost incremental learning efficiency, i.e., anti-forgetting performance on old classes and adequate compatibility with new classes. 

\subsubsection{Impact of Hyper-parameters}
\label{abla-parameters}
There are several hyper-parameters in our proposed network including $\gamma$ and $\lambda$ in Sec.~\ref{Sec-DADA} and $\Gamma$ in Sec.~\ref{Sec-DCPL}. We compare the final IL performance to reveal the impact of these hyper-parameters. Table~\ref{table-gamma} reveals the impact of $\gamma$, which can be concluded that an appropriate progressively decaying distillation intensity can boost the student network training. Table~\ref{table-lambda} shows the impact of $\lambda$. It proves the contribution of OL-D and the balance between IL-D and OL-D in the distillation process. Table~\ref{table-Aba-Gamma} shows the impact of $\Gamma$ on pseudo-labelling. The above hyper-parameters are settled based on the performance of the validation set. Please be noted that these hyper-parameters are not designed for a specific dataset, despite we only present the quantitative results on VOC 2012.	
\begin{table}[ht]
	\centering
	\begin{minipage}{0.40\textwidth}
		\centering
		\small
		\makeatletter\def\@captype{table}\makeatother
		\caption{Impact of $\gamma$ in VOC 15-1 task.} 
		\label{table-gamma}
		\setlength{\tabcolsep}{0.8mm}{
			\begin{tabular}{c|ccccccc}
				\toprule
				$\gamma$&0.5&0.6&0.7&0.8&0.9&0.95&1.0\\
				\midrule
					mIoU&64.07&65.22&66.65&67.12&\textbf{67.32}&67.28&67.19\\			
				\bottomrule
		\end{tabular}}
	\end{minipage}\quad 
	\begin{minipage}{0.35\textwidth}
		\centering
		\small
		\makeatletter\def\@captype{table}\makeatother
		\caption{Impact of $\lambda$ in VOC 15-1 task.}
		\label{table-lambda}
		\setlength{\tabcolsep}{0.8mm}{
			\begin{tabular}{c|cccccc}
				\toprule
				$\lambda$&0.5&1.0&2.0&3.0&4.0&5.0\\
				\midrule
				mIoU&66.33&67.24&\textbf{67.32}&67.28&67.13&67.15\\			
				\bottomrule
		\end{tabular}}
	\end{minipage}
	\vspace{-15pt}
\end{table}
\begin{table}[htb]
	\centering
	\caption{Impact of $\Gamma$in VOC 15-1 task. }
	\setlength{\tabcolsep}{1.2mm}{
		\begin{tabular}{c|ccccc}
			\toprule[0.4mm]
			$\Gamma$ &0.5&0.6&0.7&0.8&0.9\\
			\midrule
			mIoU&65.23&67.01&\textbf{67.32}&66.40&60.17\\		
			\bottomrule[0.4mm]
		\end{tabular}
	}
	\label{table-Aba-Gamma}
\end{table}

\subsubsection{Impact of Segmentation Model}
\label{Sec-ISM}
While most current CISS methods use DeepLabv3~\cite{DeepLabv3} as the segmentation model. However, how the segmentation model affects the CISS performance is unexplored. By using various segmentation networks, we aim to reveal the relationship between CISS efficiency and segmentation model performance. We leverage three semantic segmentation models with two kinds of encoders, i.e., DeepLabv3~\cite{DeepLabv3} with ResNet-101~\cite{ResNet}, DeepLabv3+~\cite{DeepLabv3+} with ResNet-101~\cite{ResNet} and DeepLabv3 with Swin-Transformer~\cite{SwinTH}.
The comparison among different segmentation encoders and models is conducted on VOC 15-1, ADE 100-10 and ISPRS 2-1 in Table~\ref{table-aba-segnet}. On the one hand, the impact of the decoder is explored. Compared with DeepLabv3~\cite{DeepLabv3}, DeepLabv3+~\cite{DeepLabv3+} enhances the decoder by multi-scale feature concatenation, which has been proved to be more effective on segmentation tasks. Deeplabv3+ based model achieves 4.29\% mIoU improvement on  $C^{16:20}$ and 0.54\% mIoU improvement on all classes than DeepLabv3 on VOC 15-1.  With respect to more challenging ADE 100-10 and ISPRS 2-1, it also achieves a higher mIoU rate on the initial $C^{0:100}$ and the incremental $C^{101:150}$. On the other hand, we explore the impact of the encoder on CISS performance by comparing CNN and the newfangled Transformer architecture. Specifically, we compare ResNet-101~\cite{ResNet} and  Swin-Transformer~\cite{SwinTH} in our framework. Taking ISPRS 2-1 as an example, Transformer-based architectures can bring great benefits to the final performance since its stronger feature representation ability. Therefore, it can be proved that a more robust segmentation model could boost the IL performance in CISS tasks. 

\begin{table}[h]
	\centering
	\small
	\caption{Impact of segmentation model performance in VOC 15-1, ADE 100-10 and ISPRS 2-1.} 
	\setlength{\tabcolsep}{0.45mm}{
		{\begin{tabular*}{0.46\textwidth}{@{\extracolsep{\fill}}c|l|c|ccc@{}}
				\toprule[0.4mm]
				Task &\makecell[c]{Seg.-Model} &Backbone &\multicolumn{3}{c}{\makecell[c]{mIoU (\%)}}\\
				\midrule
				&&&$C^{0:15}$ &$C^{16:20}$&all \\
				\multirow{3}{*}{\makecell[c]{VOC\\15-1}}&DeepLabv3~\cite{DeepLabv3}&ResNet-101&76.96&36.48&67.32 \\
				&DeepLabv3+~\cite{DeepLabv3+}&ResNet-101&76.32&40.77&67.86\\
				&Deeplabv3~\cite{DeepLabv3}&Swin-T&75.29&41.88& 67.34 \\
				\midrule
				&&&$C^{1:100}$ &$C^{101:150}$&all \\
				\multirow{3}{*}{\makecell[c]{ADE\\100-10}}&DeepLabv3~\cite{DeepLabv3}&ResNet-101&40.25&17.62&32.71\\
				&DeepLabv3+~\cite{DeepLabv3+}&ResNet-101&41.98&17.93&33.96\\
				&DeepLabv3\cite{DeepLabv3}&Swin-T&42.09&17.87&34.02\\
				\midrule
				&&&$C^{1:2}$ &$C^{3:5}$&all \\
				\multirow{3}{*}{\makecell[c]{ISPRS\\2-1}}&DeepLabv3~\cite{DeepLabv3}&ResNet-101&76.25&45.89&58.03\\
				&DeepLabv3+~\cite{DeepLabv3+}&ResNet-101&77.46&49.73&60.82\\
				&DeepLabv3~\cite{DeepLabv3}&Swin-T&76.42&48.35&59.58 \\
				\bottomrule[0.4mm]
		\end{tabular*}}{}}
	\label{table-aba-segnet}
\end{table}

\subsubsection{Computational Efficiency}
\begin{figure}[h]
	\centering
	\includegraphics[scale=0.52]{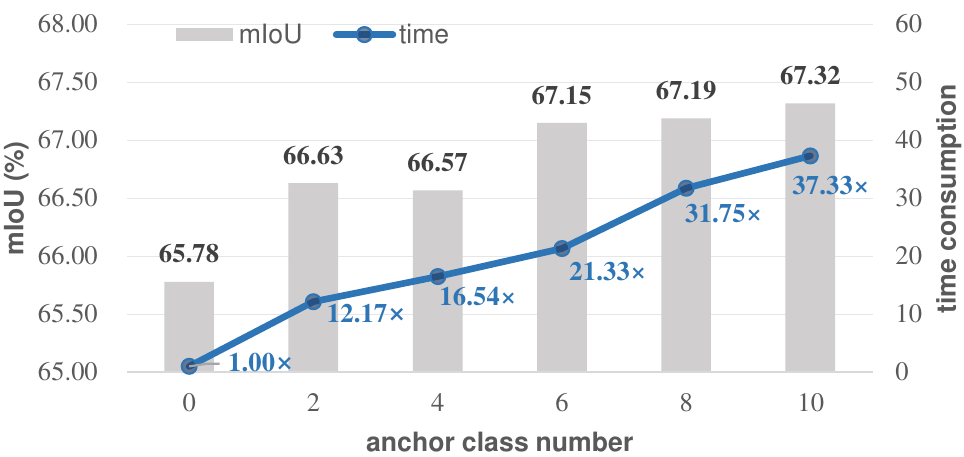}
	\caption{Impact of $\emph{anchor}$ class number in the proposed ARCL on VOC 15-1.}
	\label{fig-aba_anchor_number}
\end{figure}

Table~\ref{table-Aba-NE} reviews the training efficiency of the proposed model. In detail, we dissect module efficiency in terms of parameter, memory and floating point operations (FLOPs).  We set a baseline to represent the model that only performs CE on new classes. It is observed that DADA and ARCL, i.e., the proposed distillation mechanism and contrastive learning strategy bring large memory consumption, and ARCL encumbers training efficiency due to its region embedding filter and iterated contrastive learning process.  In Fig.~\ref{fig-aba_anchor_number}, we present the impact of \emph{anchor} class number in ARCL with mIoU evolution and time consumption in terms of VOC 15-1 task. Without ARCL, the mIoU is 65.78\% in all classes. Taking two random classes belonging to $C^{1:15}$ (by skipping the background class) as \textit{anchor} classes in every training batch, we achieve 0.95\% mIoU improvement with the price of 12.17$\times$ training time consumption. While taking ten classes as \textit{anchor} classes for contrastive learning, the model achieves the highest 67.32\% mIoU but with a cost of 37.33$\times$ time consumption to complete the training process. For a large-scale dataset such as ADE20K, setting each learned class as \textit{anchor} will bring tremendous time consumption. Thus we constrain the $anchor$ class number to ten to balance the performance and training efficiency. 

\begin{table}[h]
	\centering
	\caption{Quantitative analysis of network training efficiency in VOC 15-1 setup. Memory means GPU's memory consumption during training. The model takes 513$\times$513$\times$3 as the input size.}
	\setlength{\tabcolsep}{1mm}{
		\begin{tabular}{l|ccc}
			\toprule[0.4mm]
			\multirow{2}*{Module} &\multicolumn{3}{c}{VOC 15-1} \\
			&Param. (M) & Mem.(MB) &FLOPs (G)  \\
			\midrule
			\emph{fine tuning} &59.34&1123.22&92.89\\
			\midrule
			+DADA&74.11&1335.44&134.56 \\
			+ARCL&74.11&1351.42&137.78  \\
			\bottomrule[0.4mm]
	\end{tabular}}
	\label{table-Aba-NE}
\end{table}

\subsubsection{Impact of Class Incremental Order}
We review the impact of class incremental order to investigate the robustness of the proposed method. Taking VOC 15-1 as a representative, we set seven class orders including the ascending order and six random orders as follows:
\begin{equation}
\scriptsize
\begin{split}
\nonumber
a:\{[0,1,2,3,4,5,6,7,8,9,10,11,12,13,14,15],[16],[17],[18],[19],[20]\} \\
b:\{[0,12,9,20,7,15,8,14,16,5,19,4,1,13,2,11],[17],[3],[6],[18],[10]\} \\
c:\{[0,13,19,15,17,9,8,5,20,4,3,10,11,18,16,7],[12],[14],[6],[1],[2]\} \\
d:\{[0,15,3,2,12,14,18,20,16,11,1,19,8,10,7,17],[6],[5],[13],[9],[4]\} \\
e:\{[0,7,5,3,9,13,12,14,19,10,2,1,4,16,8,17],[15],[18],[6],[11],[20]\} \\
f:\{[0,7,13,5,11,9,2,15,12,14,3,20,1,16,4,18],[8],[6],[10],[19],[17]\} \\
g:\{[0,7,5,9,1,15,18,14,3,20,10,4,19,11,17,16],[12],[8],[6],[2],[13]\} 
\end{split}
\end{equation}

Note that Table~\ref{table-VOC2012} results from the sequential order $a$. Table~\ref{table-ClassOrders} shows the performance of IDEC on the above settings. We evaluate the mIoU on $C^{0:15}$, $C^{16:20}$ and all $C^{0:20}$. Through the standard deviation, we see that the final IL performance has slight fluctuations but is overall stable with different class orders, validating the robustness of the model. 
In addition, we display the detailed performance in VOC 15-5, 15-1, 5-3 and 10-1 for each class in Table~\ref{table-DetailClass} under the sequential class order \textit{a}.
\begin{table}[h]
	\centering
	\caption{Impact of class incremental order on VOC 15-1 task.}
	\setlength{\tabcolsep}{3.0mm}{
		\begin{tabular}{c|cc|c}
			\toprule[0.4mm]
			order&0-15&16-20&all\\
			\midrule
			a&76.96&36.48&67.32\\
			b&75.82&41.22&67.58\\
			c&74.73&28.61&63.75\\
			d&74.20&36.53&65.23\\
			e&73.05&37.49&64.58\\
			f&68.87&40.00&62.00\\
			g&74.79&39.46&66.38 \\
			\midrule
			avg.$\pm$std.&-&-&65.26$\pm$1.86\\
			\bottomrule[0.4mm]
	\end{tabular}}
	\label{table-ClassOrders}
\end{table}
\begin{table*}[htbp]
	\centering
	\caption{Class-specific performance on VOC 15-1 after the final IL step in IoU (\%). The results come from the proposed method.}
	\setlength{\tabcolsep}{0.8mm}{
		\begin{tabular}{c|ccccccccccccccccccccc}
			\toprule[0.4mm]
			class&\begin{sideways}bg\end{sideways}&\begin{sideways}aeroplane\end{sideways}&\begin{sideways}bicycle\end{sideways}&\begin{sideways}bird\end{sideways}&\begin{sideways}boat\end{sideways}&\begin{sideways}bottle\end{sideways}&\begin{sideways}bus\end{sideways}&\begin{sideways}car\end{sideways}&\begin{sideways}cat\end{sideways}&\begin{sideways}chair\end{sideways}&\begin{sideways}cow\end{sideways}&\begin{sideways}diningtable\end{sideways}&\begin{sideways}dog\end{sideways}&\begin{sideways}horse\end{sideways}&\begin{sideways}motorbike\end{sideways}&\begin{sideways}person\end{sideways}&\begin{sideways}pottedplant\end{sideways}&\begin{sideways}sheep\end{sideways}&\begin{sideways}sofa\end{sideways}&\begin{sideways}train\end{sideways}&\begin{sideways}monitor\end{sideways}\\
			\midrule
			15-5&91.29&87.63&39.83&90.89&68.89&77.01&90.25&88.55&93.67&35.28&85.50&57.01&88.49&82.46&85.85&85.46&32.70&68.81&29.15&72.07&56.47\\
			15-1&89.45&86.73&40.24&88.58&66.43&75.62&87.56&88.43&92.84&36.00&75.63&59.33&89.13&83.48&86.55&85.43&29.28&48.64&23.55&47.31&33.61\\
			5-3&88.09&66.06&33.01&84.29&57.14&73.70&64.68&67.91&75.34&13.41&42.45&37.73&65.36&41.89&63.29&76.09&27.78&40.47&18.85&49.25&50.14\\
			10-1&87.62&78.21&36.88&86.32&60.18&72.12&83.68&83.83&88.92&32.71&67.70&30.47&74.56&46.57&65.51&78.33&22.00&39.27&24.39&39.31&42.54\\
			\bottomrule[0.4mm]
	\end{tabular}}
	\label{table-DetailClass}
\end{table*}

\subsubsection{Failure Example and Analysis}
\begin{figure}[h]
	\centering
	\includegraphics[scale=0.45]{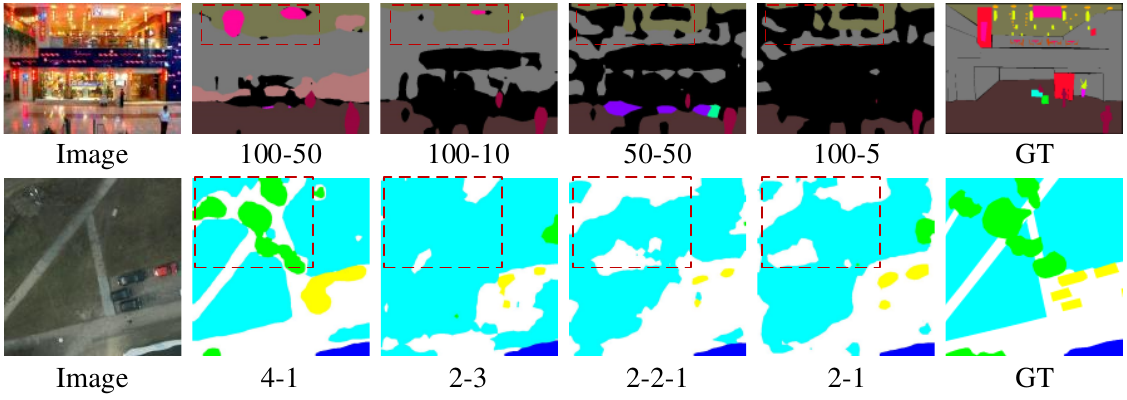}
	\caption{Failure example.}
	\label{fig-failure_example}
\end{figure}

We demonstrate the effectiveness of the proposed method in both natural and remote-sensing scenes aforesaid. Taking a closer look at the prediction results, there are some flaws including 1) Complex scenes: as shown in line 1 in Fig~\ref{fig-failure_example}, the segmentation model fails to recognize the complex semantic content,  aggravating the catastrophic forgetting. 2) Hard examples: in Fig~\ref{fig-failure_example} line 2, class \emph{tree} and \emph{low vegetation} possess approximate colour and texture, causing pixel misclassification in multi-step tasks. We believe these obstacles can be alleviated by employing a stronger semantic segmentation model.

\section{Conclusion}
In this paper, we propose to addressing class incremental semantic segmentation without data replay. We reveal the associativity of catastrophic forgetting and semantic drift and review the progress in this field. To resolve these two problems, we make three main contributions: knowledge distillation to alleviate the former, and contrastive learning in asymmetric latent space to increase the discriminability between learned and incremental classes for the latter. Therewith a customized class-specific pseudo-labelling strategy is used to boost learning efficiency. They prove their effectiveness on various CISS tasks (VOC 2012, ADE20K, ISPRS) and settings (12 in total). And the most conspicuous point is we achieve robust CISS performance especially in multi-step tasks. We also demonstrate the anti-forgetting and model evolving can be non-antagonistic but concordant through enhancing the ability to differentiate old and new classes in the latent space. However, a crack in our contrastive learning strategy is that it requires vast time computation for training. If not comparing all classes at each incremental step may alleviate this problem. Our future work will focus on exploring human-like anti-forgetting mechanisms and adapting our method to more scenes like open-set situation, as well as few-shot/zero-shot CISS study.

\ifCLASSOPTIONcompsoc
  \section*{Acknowledgments}
\else
\section*{Acknowledgment}
\fi

This work was supported by the National Natural Science Foundation of China under Grant 62271018.

\ifCLASSOPTIONcaptionsoff
  \newpage
\fi

{\small
	\bibliographystyle{IEEEtran}
	\bibliography{egbib}
}

\end{document}